\documentclass[hyperref]{article}
\usepackage{lineno,hyperref}
\modulolinenumbers[5]

\usepackage[a4paper,left=18mm,right=18mm,top=15mm,bottom=15mm]{geometry}
\usepackage{amsmath,amssymb,amsfonts}
\usepackage{graphicx}
\usepackage{textcomp}
\usepackage{xcolor}
\usepackage{diagbox}
\usepackage{subfigure}
\usepackage{booktabs}
\usepackage{multirow}
\usepackage{makecell}
\usepackage{color}
\usepackage{float}
\usepackage{url}
\usepackage{booktabs}
\usepackage{cleveref}
\usepackage{mathrsfs}

\usepackage{bm}
\usepackage{textcomp} 
\usepackage{bbding} 
\usepackage{stfloats} 
\usepackage{diagbox} 

\usepackage[linesnumbered,ruled,boxed]{algorithm2e}
\usepackage{algpseudocode}

\usepackage{authblk}

\def\BibTeX{{\rm B\kern-.05em{\sc i\kern-.025em b}\kern-.08em
    T\kern-.1667em\lower.7ex\hbox{E}\kern-.125emX}}



\begin{document}


\title{\textbf{Addressing modern and practical challenges in machine learning: A survey of online federated and transfer learning}}
%

%
\author[1]{Shuang Dai\thanks{sd19628@essex.ac.uk}}
\author[1]{Fanlin Meng\thanks{fanlin.meng@essex.ac.uk}}
\affil[1]{Department of Mathematical Sciences, University of Essex, Colchester, UK}
\date{January 2022}

\maketitle



\begin{abstract}
  Online federated learning (OFL) and online transfer learning (OTL) are two collaborative paradigms for overcoming modern machine learning challenges such as data silos, streaming data, and data security. This survey explored OFL and OTL throughout their major evolutionary routes to enhance understanding of online federated and transfer learning. Besides, practical aspects of popular datasets and cutting-edge applications for online federated and transfer learning are highlighted in this work. Furthermore, this survey provides insight into potential future research areas and aims to serve as a resource for professionals developing online federated and transfer learning frameworks.\\
  \textbf{\textit{keywords:}} online transfer learning, online federated learning, online learning, federated transfer learning, privacy-preserving
\end{abstract}

\section{Introduction}
Recent advancements in machine learning have propelled the broad utilization of smart technologies, particularly the Internet of Things (IoT). Worldwide, IoT devices are expected to nearly triple from 8.74 billion in 2020 to over 25 billion in 2030 \cite{NumberofIoT}. On the one hand, massive data collected from IoT devices are critical for constructing robust machine learning models, which have created a wealth of chances for growing innovations in the era of big data. Besides, real-world machine learning advances have relied on the availability of enormous amounts of well-labeled data, such as ImgNet \cite{deng2009imagenet} and Alpha Zero \cite{holcomb2018overview}. On the other hand, big data, which are characterized by high volume, high velocity, and high diversity \cite{manyika2011big}, cannot be utilized directly as high-quality ready inputs, posing many obstacles to the development of data-driven real-world machine learning systems.

The challenges of developing data-driven machine learning systems in the era of big data are distinct from those of classic theoretical frameworks, owing to the features of big data and the restrictions placed by data regulations and laws, such as the new General Data Protection Regulation (GDPR) \cite{GDPR}. These distinctions have important effects on the assumptions and performance measures underlying the design of data-driven machine learning systems and may stimulate the development of more innovative and practical machine learning algorithms. Therefore, in the following, we will first identify the modern challenges in real-world machine learning and then give an overview and highlight the contributions of this survey.

\subsection{Modern challenges in real-world machine learning}
Machine learning has been widely applied in various real-world applications and achieved satisfactory performance. Generally, a good machine learning model requires plentiful training data and a well-designed model. Therefore, we identify significant modern challenges in the era of big data and discuss their impact on developing real-world machine learning models at both the data and model levels.

From the data standpoint, high-quality datasets can provide more comprehensive information essential for building an effective machine learning model. However, in real-world machine learning applications, data may not be stored in a centralized location, and exhibit statistical disparities \cite{li2019convergence}, referred to as \textit{data silos}. Medical records, for example, are private and stored in isolated medical facilities; several facilities may only contain unlabeled data whereas others may contain only a few labeled records. Besides, \textit{data labeling} is prohibitively expensive, especially in fields that require human skill and domain expertise, such as the medical sector. Therefore, the lack of labeled data is another obstacle to the development of real-world machine learning since model performance is highly dependent on labeled data \cite{deng2009imagenet}. Also, data collection has become increasingly challenging from a legislative standpoint, which is referred to as \textit{data governance}. For example, the GDPR \cite{GDPR} has several provisions that safeguard user privacy and restrict companies from transferring data without explicit user consent. Moreover, the \textit{real-time data} collected by IoTs allow a more effective allocation of resources and pose additional difficulties for conventional offline machine learning frameworks that rely on pre-given training data. For instance, real-time traffic data on road conditions are gathered and analyzed to improve traffic management in smart cities, which necessitates a dynamic machine learning framework capable of handling training samples that arrive in an online manner \cite{jeong2013supervised}.

From the model perspective, a well-designed model can make effective inferences and fit for personalized requirements of various tasks. However, the non-independent and identically distributed (non-IID) distribution of data in real-world complicates the training of a single model that can work effectively for all tasks. For instance, when the next-word prediction task is applied to a certain phrase, it should suggest a response tailored to each local user. Different local users label the same data differently, necessitating the development of customized models \cite{kulkarni2020survey}. As a result, \textit{model personalization} is increasingly popular to meet the diverse needs of various users. Another current challenge in real-world machine learning is rapidly inferring a high-performance model for new users and effectively updating existing models, i.e., \textit{constructing models effectively}. Considering a distributed system as an example; Conventional machine learning models, which are based on a pre-given dataset, need to be retrained whenever new users join the system. This will lead to the wasting bandwidth and computing resources.

Various solutions have been proposed to address the aforementioned challenges, including online transfer learning (OTL) \cite{OTL10}, and online federated learning (OFL) \cite{OFL6}. OTL and OFL take transfer learning (TL) \cite{pan2009survey} and federated learning (FL) \cite{yang2019federated} in the online context, allowing these advanced methods to process online big data efficiently. OTL aims to leverage knowledge from source domains to develop an online model for the target domain, addressing the challenges associated with the lack of well-labeled data for efficient online prediction for sequentially arriving data in the target domain. While OTL addresses the problem of data labeling in the online context, it still requires central access to data in both the source and target domains, which may violate data privacy and security standards in the big data era. On the other hand, OFL focuses on training a central model that makes use of real-time data generated by multiple distributed local devices without violating data privacy regulations. During each training round, only updated parameters of each local model are sent to the central model, ensuring the central model performance while maintaining privacy.

\subsection{Overview of this survey}

The purpose of this paper is to provide a detailed survey of various methods for addressing modern machine learning challenges, focusing on OFL and OTL.

\begin{figure}[!htbp]
\centering
\includegraphics[width=0.5\textwidth]{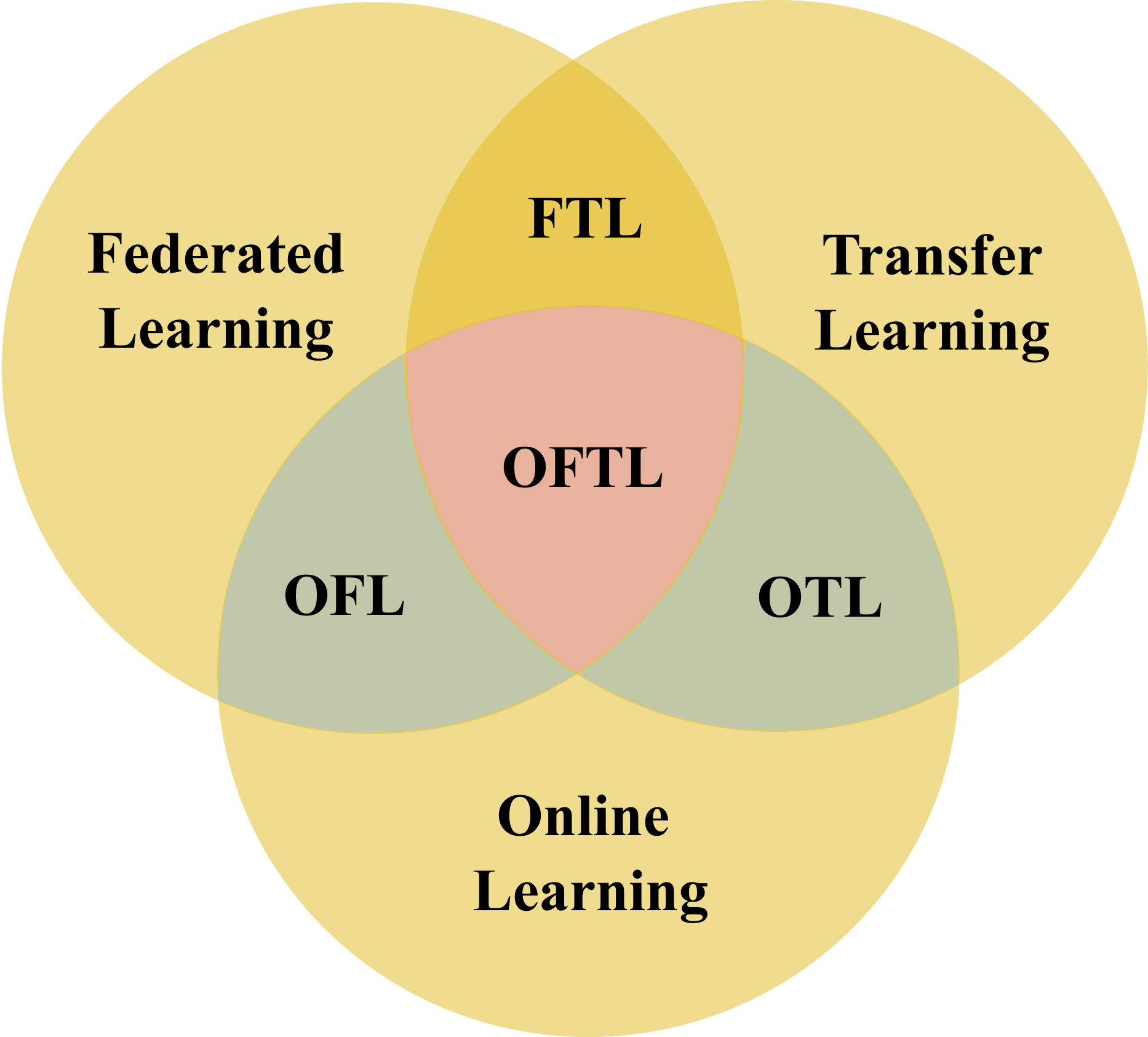}
\caption{Blueprint of our survey. FTL: Federated transfer learning; OFTL: Online federated transfer learning.}
\label{Fig1}
\end{figure}

Fig. \ref{Fig1} illustrates the blueprint of this survey. The green areas are our main emphasis, whereas existing surveys only concentrate on the yellow areas. The red section is one of the critical future paths we suggested for further investigation. We consider federated and transfer learning in online scenarios: OTL is not studied by traditional learning types of TL \cite{pan2009survey} \cite{weiss2016survey}, i.e., transductive, inductive, and unsupervised TL. Instead, we discuss OTL from two viewpoints: domain-based OTL and task-based OTL. Furthermore,
we review OFL from three aspects: statistical heterogeneity, system heterogeneity, and privacy guarantees, highlighting the most significant challenges. The main contributions of our work are summarized as follows:

\begin{itemize}
  \item To the best of our knowledge, this is the \textit{first} survey to present the recent advances in OTL and OFL existing studies, and pointed out potential future research directions. This survey aims to serve as a resource for researchers and practitioners developing online federated and transfer learning frameworks.
  \item We provide definitions for OTL and OFL, as well as new viewpoints on them. Additionally, we describe in detail of recent advances in online federated and transfer learning, and highlight the connections between different methods.
  \item We summarize popular datasets and cutting-edge applications of OTL and OFL, and discuss practical considerations and provide insights into potential future research directions.
\end{itemize}

The remainder of this survey is structured as follows. Section \ref{2} reviews and reports on related works, which provides necessary backgrounds of OTL and OFL. Then, recent advances in OTL and OFL are reviewed in Section \ref{3} and \ref{4}, respectively. Practical considerations in datasets and applications of OTL and OFL are summarized and presented in Section \ref{5}. In Section \ref{6}, we conclude this survey and discuss future research directions.

\section{Related work}\label{2}
We review related work on OTL and OFL in this part, including TL, FL, FTL, and OL. Moreover, we summarize the implementation scenarios of these methods and point out the existing challenges associated with them.

\subsection{Transfer learning}

\begin{figure}[!htbp]
\centering
\includegraphics[width=0.5\textwidth]{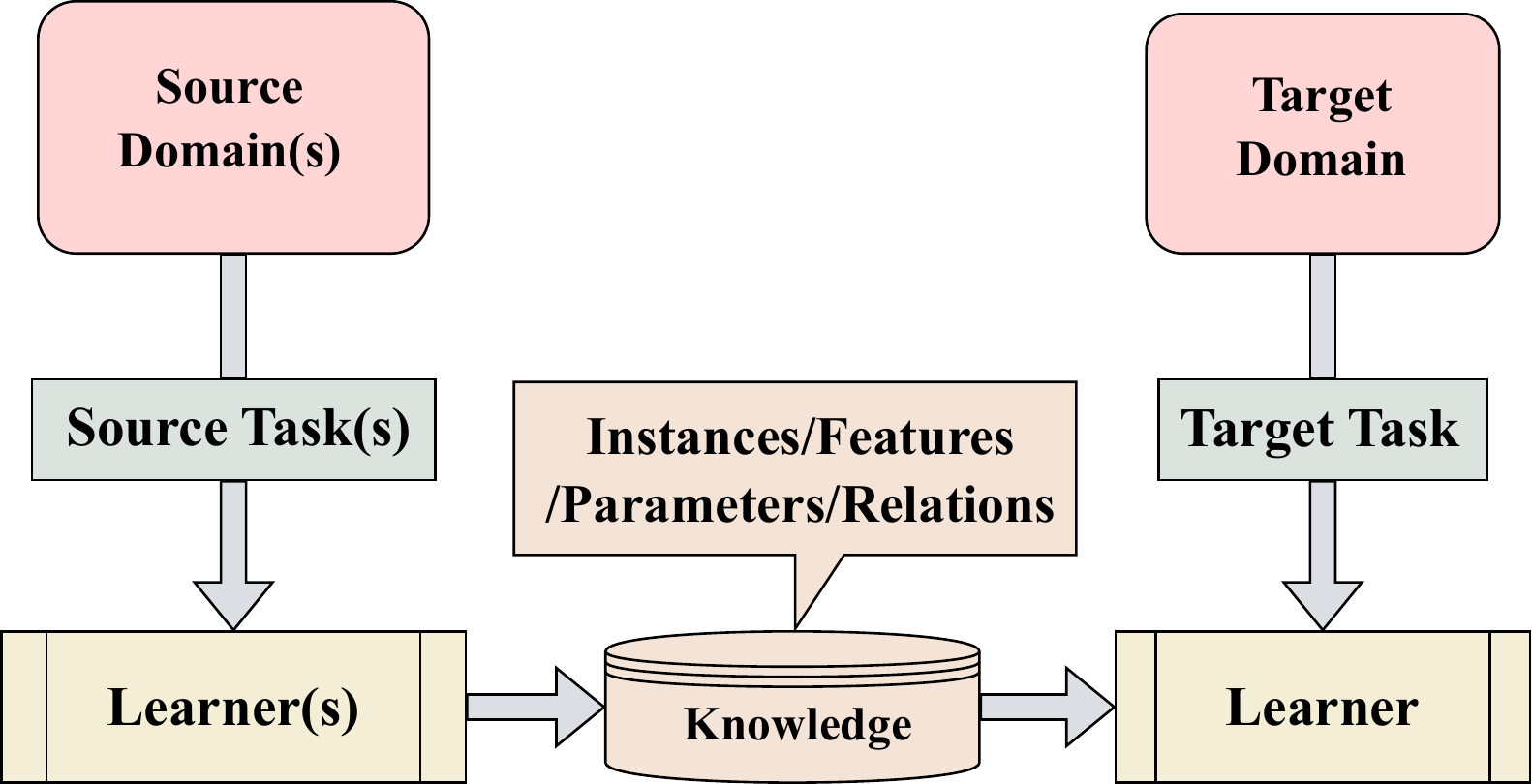}
\caption{Transfer learning process.}
\label{Fig3}
\end{figure}

The assumption made by most of the traditional machine learning algorithms is that the training and test data have the same distribution and feature space. However, this assumption does not hold in the majority of real-world scenarios. Furthermore, traditional machine learning has been hampered by a lack of adequately labeled training data and mismatched computing capability. TL \cite{pan2009survey} was proposed to address these challenges by leveraging knowledge from a single or multiple source domains to enhance a training task in the target domain (Fig. \ref{Fig3}). The knowledge transferred could be instances from source domains \cite{deca31, deca32, deca79}, shared features from source domains and the target domain \cite{deca18, deca19, deca22}, parameters from the trained learners of source domains \cite{deca51, deca53}, or relations between source domains and the target domain \cite{deca82}.

According to different implementation scenarios, TL can be categorized as single source TL and multiple sources TL. Single source TL refers to transferring knowledge from single source domain \cite{fenlei32} whereas the multiple sources TL utilizes several source domains to transfer the knowledge \cite{fenlei46, fenlei49}. Moreover, different TL techniques have been proposed to handle similar or different data structures between the source and target domains, i.e., homogeneous and heterogeneous TL \cite{day2017survey, zhou2019deep}.

According to different label settings, a variety of TL methods have been proposed and can be classified into three major categories, i.e., transductive, inductive, and unsupervised TL \cite{pan2009survey}.

Inductive TL is used when well-labeled target domain data is available, and there are different tasks in the source and target domains. TrAdaBoost \cite{deca42} is a well-known inductive TL technique that extracts valuable information from the source domain by re-weighting predicted instances in both the source and target domains. However, this method only utilized a single source domain, and the extracted information may not be sufficient for the training task in the target domain. As a result, \cite{deca45, deca34} combined the transfer task with multiple source domains, which enhanced the training performance of the target model. Unlike \cite{deca42}, which retains only one base learner and discards the others, \cite{deca76} argues that all base learners are useful, based on the theory that older learners can represent the major distributions of instances, while newer learners can provide detailed information about subsequent iterations.

Transductive TL is used when the source domain data is labeled, but the target domain data is unlabeled, and both the source and target domains have the same task. Domain adaptation is the most well-known subfield of transductive TL \cite{niu2020decade}, which attempts to minimize the marginal distribution gap between the source and target domains. \cite{deca31} proposed instance selection and weighting methods based on PU learning for identifying the examples that can improve the training task from the source domain. However, this method was hampered by the difficulty of dealing with high-dimensional distributions. \cite{deca32} provided a solution to this issue by using the logistic approximation to adapt the high-dimensional data from the source domain to the target domain.

In real-world situations, both the source and target domains may be insufficient in well-labeled data, which cannot be addressed using the TL techniques discussed previously. To address such an issue, unsupervised TL was introduced. \cite{deca54} proposed transferred discriminative analysis (TDA), a method that leverages related prior knowledge from the source domain to produce class labels for unlabeled target data. Although unsupervised learning approaches solve the transfer task from a more practical perspective, it only received little attention from researchers over the last decade.

\subsection{Federated learning}
\begin{figure}[!htbp]
\centering
\includegraphics[width=0.5\textwidth]{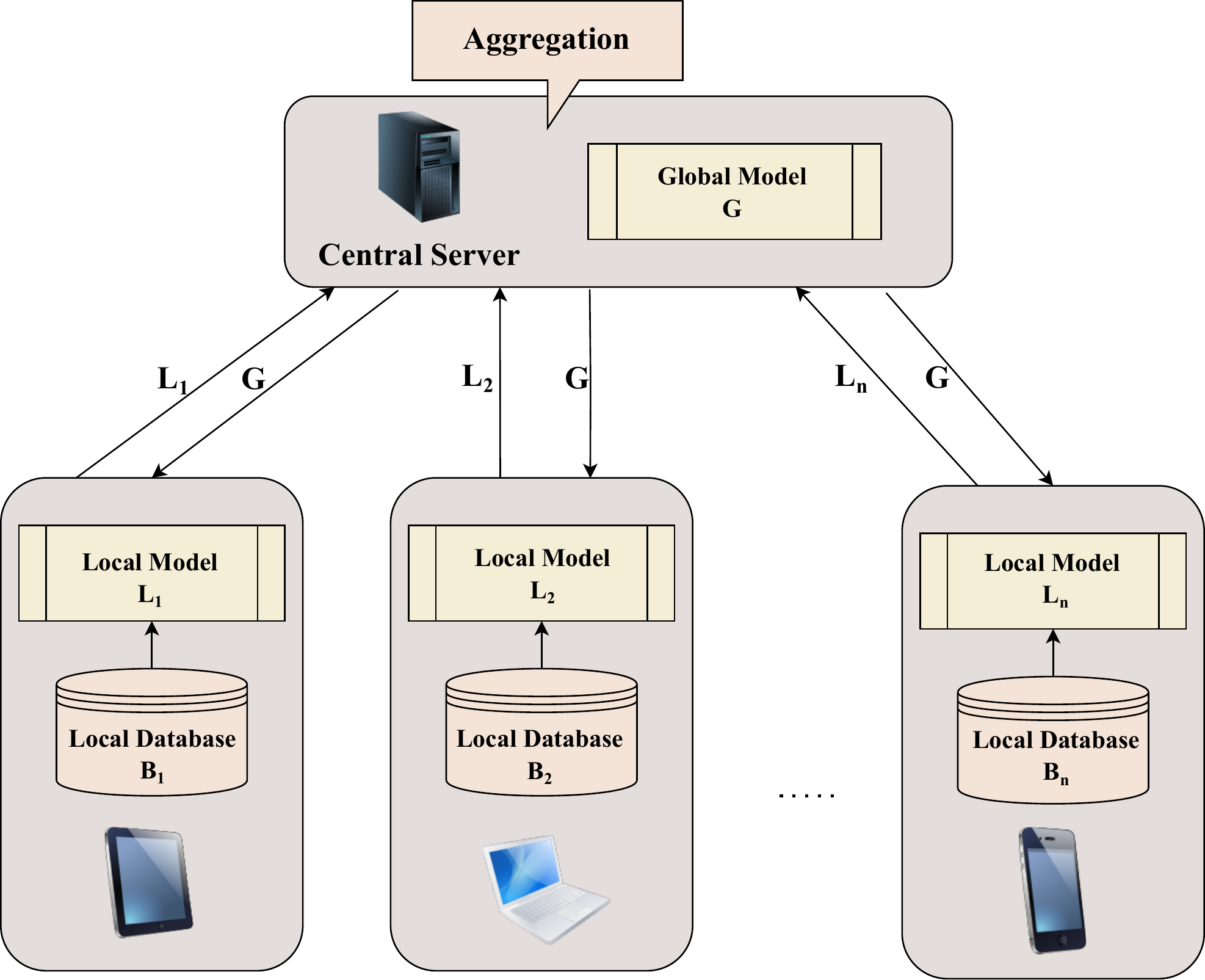}
\caption{Federated learning process.}
\label{Fig4}
\end{figure}

IoTs, such as smart healthcare devices and smart meters, continuously collect massive data. While models trained by the aggregated data from these applications enable efficient management of smart cities applications, the process is complicated by a variety of legal constraints. FL has been proposed in this context for training a global model from data stored on multiple distributed devices, with only intermediate updates periodically sent to a central server \cite{yang2019federated}. A typical FL paradigm is illustrated in Fig. \ref{Fig4}, in which the central server first shares the initial model parameters with all local clients, and then each client trains the local model and updates the parameters to the central server. The central server then uses the uploaded parameters to update the global model and broadcasts the revised global parameters to the local clients. The above processes are repeated continuously to ensure that the global model is updated and optimized across all local clients.

FL can be categorised into horizontal FL, vertical FL, and FTL, depending on how data is distributed among different devices in the sample and the feature space. Besides, since FTL is known as a novel combination of TL and FL, we will discuss this technique in more detail in chapter (\ref{FTL}).


Horizontal federated learning (HFL) (Fig. \ref{Fig4}) refers to the situation in which data from distributed devices share the same feature space but differ in samples. Google pioneered HFL by utilising data distributed across many local Android devices to forecast text input without breaking privacy regulations \cite{mcmahan2017communication}. \cite{gao2019hhhfl} then developed a hierarchical heterogeneous HFL architecture to extend HFL into heterogeneous environments, effectively addressing the issue of inadequate labeled data in local source devices. \cite{FML9} designed a secure aggregation scheme to further enhance the privacy of aggregated intermediate updates based on \cite{mcmahan2017communication}. Furthermore, researches \cite{smith2017federated, FML36} have been proposed to address the high cost of communication in the conventional FL framework.

\begin{figure}[!htbp]
\centering
\includegraphics[width=1\textwidth]{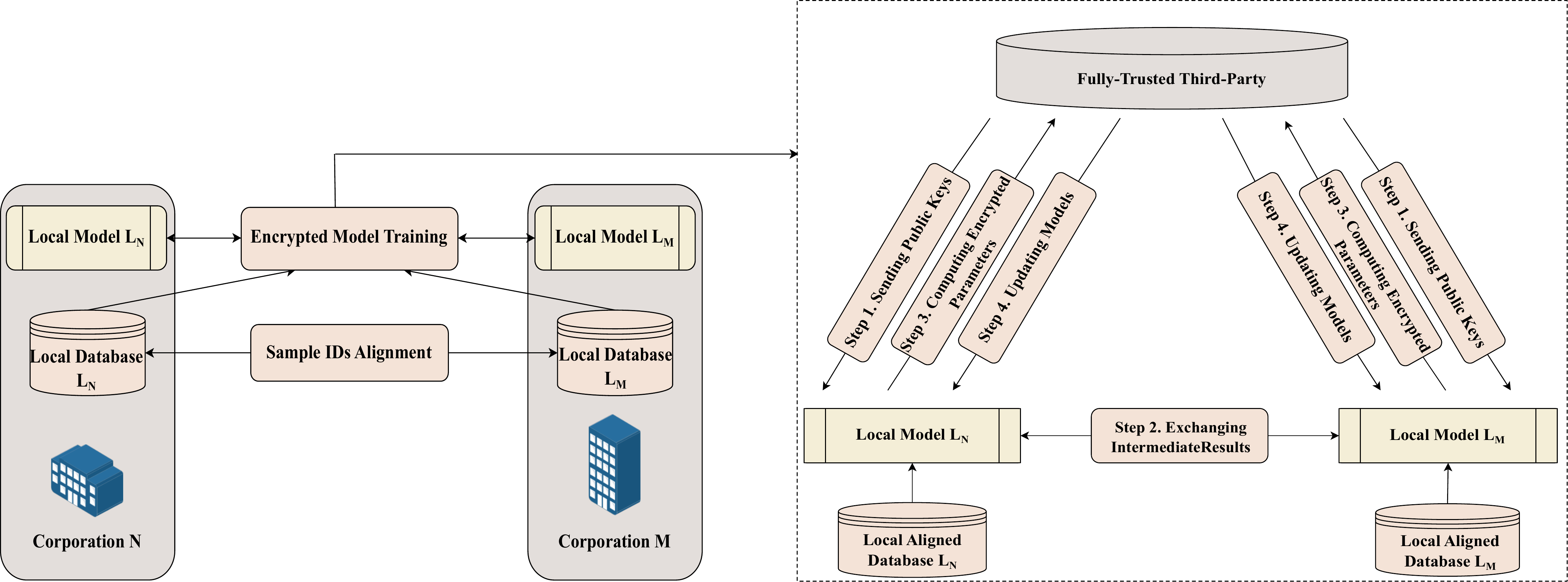}
\caption{Vertical Federated learning process.}
\label{Fig5}
\end{figure}

Vertical federated learning (VFL) was proposed on the premise that heterogeneous data from various devices share common sample IDs but have distinct feature spaces, and thus VFL focuses on the correlation between devices from different sectors. In a typical VFL process, data with common sample IDs should be retrieved and used to train the machine learning model (subfigure on the right in Fig. \ref{Fig5}). VFL is more difficult to implement than HFL since it needs encrypted user-ID alignment algorithms \cite{FML56} to extract common entities \cite{yang2019federated} and also requires the authentication of a fully trusted third-party. To overcome these obstacles and simplify VFL, \cite{yang2019parallel} developed a framework that eliminates the need for a third-party coordinator, and this framework has been shown to be efficient and scalable. Although VFL is capable of handling heterogeneous data from a variety of domains, the majority of VFL techniques rely on statistical models such as logistic regression rather than sophisticated machine learning frameworks, indicating that this field still needs enormous efforts.

Apart from data distribution, FL can be classified in various ways. Based on network topology, FL can be classified into centralized FL and peer-to-peer (P2P) FL \cite{2021AS35, 2021AS46}. Centralized FL generally relies on a centralized server to aggregate and broadcast the updated parameters. In contrast to centralized FL, P2P FL does not rely on a global server for local model updates, and instead exchanges parameters between neighbors directly. Based on data availability, FL can also be classified into cross-silo FL, and cross-device FL \cite{kairouz2019advances}. Since almost every local client in the cross-silo FL is considered indexed and available for updating at any time, this framework is well-suited for scenarios involving a small number of local clients, where the siloed data are from geo-distributed data centers (e.g., local banks or medical centers) instead of the large number of distributed edge nodes (e.g., smartphones or laptops). On the other hand, cross-device FL is used when there are a large number of participants and the local clients are not constantly available and reliable. To compensate for the unreliability of local clients, the cross-device FL often employs resource allocation techniques \cite{2021AS61} and incentive mechanisms \cite{2021AS64} to improve the overall performance of the FL framework.

\subsection{Federated transfer learning} \label{FTL}
\begin{figure}[!htbp]
\centering
\includegraphics[width=0.5\textwidth]{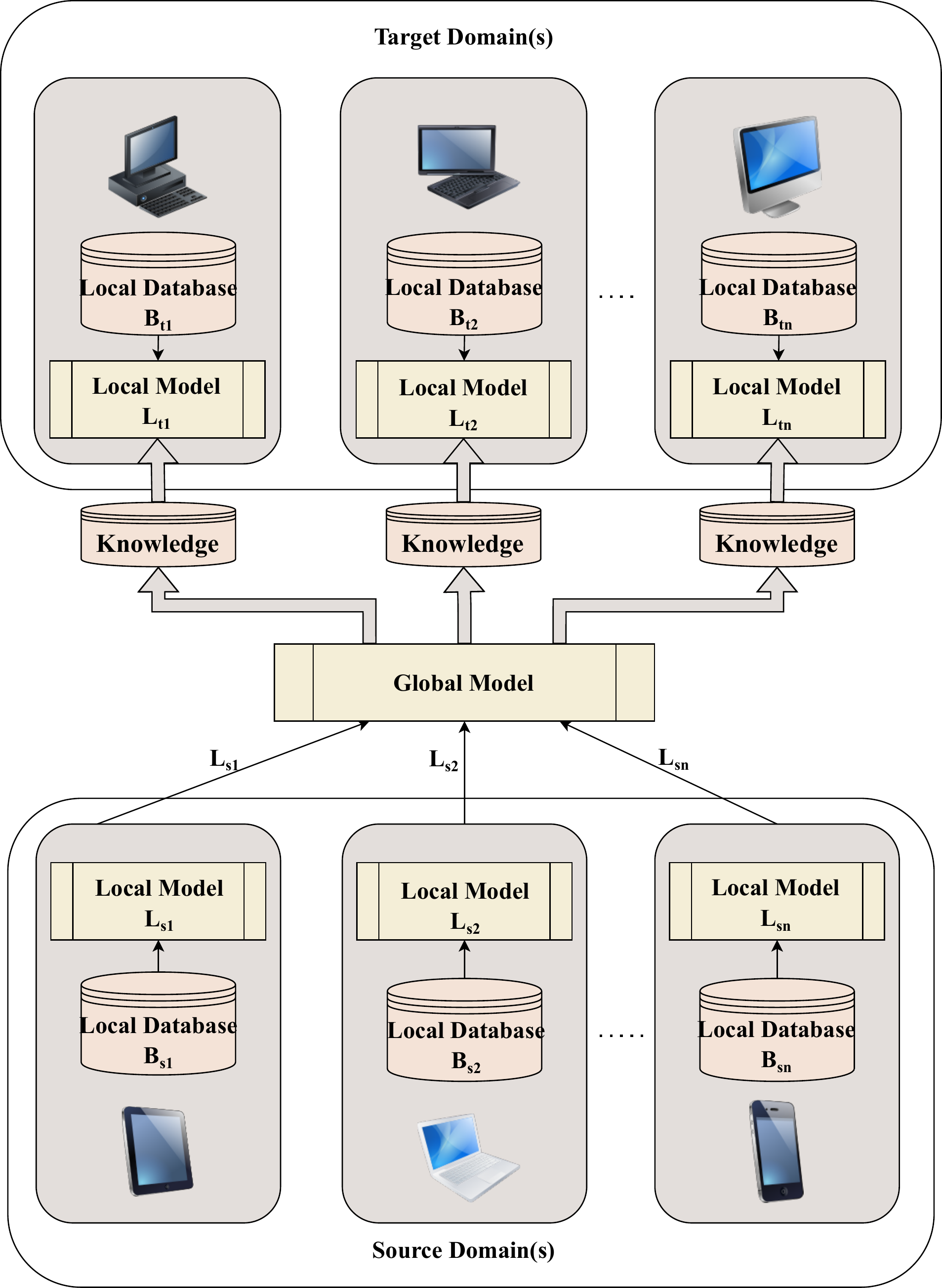}
\caption{Federated transfer learning process.}
\label{Fig6}
\end{figure}

Different from HFL and VFL, FTL \cite{liu2020secure} refers to the situation in which data across multiple devices differ in terms of both feature spaces and sample IDs and is regarded as a significant extension of traditional FL frameworks \cite{yang2019federated}. Rather than limiting conditions to sharing only matching data (i.e. data with overlapped feature spaces or sample IDs), FTL enables users to leverage large datasets with well-trained machine learning model parameters to meet their specific needs \cite{mothukuri2021survey}, and Fig. \ref{Fig6} depicts the general process of FTL. The use of TL in FL systems addresses the issue of lack of well-labeled data in the source devices and enables various sectors to securely and privately train more personalized local models. It is worth pointing out that while TL and FL are natural complements, relatively few studies have considered the FTL framework.

Similar to conventional FL methods, the impediment to FTL development is training data in heterogeneous settings, which is made even more difficult by the restrictive assumption of FTL application scenarios. \cite{dimitriadis2020federated} developed a dynamic gradient aggregation algorithm to address this problem by regularising the local updated gradient. To enable FTL in heterogeneous intelligent manufacturing applications, \cite{kevin2021federated} utilized pre-built models from a variety of smart environments as the central source domain, and the central server then chose the best model to broadcast based on the similarity between the central source models and the local target models. Thus, each heterogeneous local device will conduct TL to acquire application-specific models. Additionally, communication efficiency is another concern in FTL. \cite{sharma2019secure} adopted secret sharing (SS) to improve the communication efficiency and also the privacy level of FTL, which allowed servers to be malicious.

FTL has received growing interests in real-world applications, such as smart healthcare \cite{chen2020fedhealth}, traffic monitoring \cite{liang2019federated}, smart energy \cite{zhang2021fednilm}, and image analysis \cite{yang2020fedsteg}. The majority of currently available FTL systems are based on deep learning architectures \cite{chen2020fedhealth, zhang2021fednilm, yang2020fedsteg, kevin2021federated} that usually freeze the base layers of the global model and retrain the fully-connected layer on local devices. \cite{chen2020fedhealth} performed human activity recognition via FTL, which replaced one of the fully-connected layers with a correlation alignment layer to facilitate domain adaptation. The highly transferable features in the low-level layers and the specific features capturing ability in the high-level layers of the deep network made FTL with deep learning architectures efficient \cite{yosinski2014transferable}.

\subsection{Online learning}
OL is a machine learning paradigm for real-time data in which a classifier attempts to learn and update the best predictor for future data using feedback from the sequence data at each step. In comparison to the optimal model in foresight, the primary goal of OL is to minimize cumulative error across the whole data sequence \cite{hoi2014libol}. Compared to conventional batch learning algorithms, which require pre-given training data, OL is generally more effective and scalable when dealing with large-scale real-world machine learning problems involving data of varying quantity and velocity.

OL has been extensively investigated for many years \cite{cesa2006prediction} \cite{hoi2014libol}. There are two fundamental types of OL algorithms: first-order OL and second-order OL. \cite{hoi2014libol}. The Perceptron \cite{rosenblatt1958perceptron} \cite{novikoff1963convergence} is one of the earliest first-order OL algorithms, relying on gradient feedback to update a linear classifier whenever a new sample is misclassified. Passive-Aggressive (PA) \cite{crammer2006online} was introduced as a family of first-order OL algorithms based on margin-based learning. It updates the model when the classification confidence of a new sample falls below a predefined threshold. Moreover, online gradient descent \cite{zinkevich2003online} \cite{bartlett2009adaptive} \cite{dekel2012optimal} was proposed to model the OL as an online convex optimization problem.

The misclassified instances are retained as support vectors (SVs) in standard OL algorithms (e.g., Perceptron and PA). Despite their solid theoretical guarantees and efficient functioning, a fundamental issue is that the increasing number of SVs over time may result in the constant increase in computation overheads. To overcome this challenge, \cite{dekel2008forgetron} discarded the oldest SVs assuming that they were less representative of the data stream distributions. Additionally, \cite{zhao2012fast} presented bounded online gradient descent (BOGD) to constrain the amount of SVs that fall below a threshold.

Unlike first-order OL algorithms, which maximize convergence by utilizing only the first-order derivative information of the gradient, second-order OL algorithms maximize convergence by utilizing both the first-order and second-order information. The second-order Perceptron algorithm \cite{cesa2005second} was designed to examine the geometric properties of data. In order to capture second-order information about the confidence level of the features, the confidence weighted (CW) algorithm \cite{dredze2008confidence} was developed to manage the update process of the classifier. Second-order OL require exponential space and time for updates, and the sketched online Newton (SON) \cite{luo2016efficient} was introduced to address this issue. The SON is an enhanced version of the online Newton step with a linear running time in dimension and sketch size, allowing for dramatic improvements in second-order learning efficiency.

\subsection{Frontier implementation scenarios and inter-connections of TL, FL, FTL, and OL}\label{scenarios}

TL, FL, FTL, and OL are all innovative approaches built on standard machine learning techniques to address modern challenges in real-world applications. In this subsection, we will outline their implementation scenarios to investigate the underlying relationship between them and discuss the existing challenges to emphasize the significance of our survey. Table \ref{Tab1} compares the implementation scenarios of traditional machine learning, TL, FL, FTL, and OL, which can be used as a guide to assist professionals in selecting the most appropriate methods to apply to specific real-world problems.

\renewcommand\arraystretch{1.4}
\begin{table*}[!htb]
\caption{Frontier implementation scenarios of different techniques}
\resizebox{\textwidth}{14mm}{
\begin{tabular}{lccclcccc}
\hline
\multicolumn{1}{c}{\multirow{2}{*}{}} & \multirow{2}{*}{\textbf{Decentralization}} & \multicolumn{3}{c}{\textbf{Heterogeneity}}                                                                     & \multirow{2}{*}{\textbf{\begin{tabular}[c]{@{}c@{}}Inadequate Well-labeled \\ Data\end{tabular}}}
& \multirow{2}{*}{\textbf{Privacy-Preserving}} &
 \multirow{2}{*}{\textbf{\begin{tabular}[c]{@{}c@{}}Client-Side \\ Personalization\end{tabular}}}
 & \multirow{2}{*}{\textbf{\begin{tabular}[c]{@{}c@{}}Real-time \\ Data\end{tabular}}} \\ \cline{3-5}
\multicolumn{1}{c}{}                  &                                            & \multicolumn{1}{l}{\textbf{Cross-Modality}} & \multicolumn{1}{l}{\textbf{Cross-Model}} & \textbf{Cross-System} &                                                                                                 &                                              &                                                                                                  &                                                                                                                                                                                             \\ \hline
\textbf{Traditional Machine Learning} & \XSolid                                    & \XSolid                                     & \XSolid                                  & \XSolid               & \XSolid                                                                                         & \XSolid                                                                                        & \XSolid                                                                                 & \XSolid                                                                                             \\
\textbf{Transfer Learning}            & \XSolid                                    & \Checkmark                                  & \Checkmark                               & \XSolid               & \Checkmark                                                                                      & \XSolid                                                                                       & \Checkmark                                                                                 & \XSolid                                                                                          \\
\textbf{Federated Learning}           & \Checkmark                                 & \Checkmark                                  & \Checkmark                               & \Checkmark            & \Checkmark                                                                                      & \Checkmark                                                                                     & \XSolid                                                                                 & \XSolid                                                                                             \\
\textbf{Federated Transfer Learning}  & \Checkmark                                 & \Checkmark                                  & \Checkmark                               & \Checkmark            & \Checkmark                                                                                      & \Checkmark                                                                                      & \Checkmark                                                                                 & \XSolid                                                                                          \\
\textbf{Online Learning}            & \XSolid                                    & \XSolid                                     & \XSolid                                  & \XSolid               & \XSolid                                                                                         & \XSolid                                                                                         & \XSolid                                                                                 & \Checkmark                                                                                          \\
\hline
\end{tabular}}
\label{Tab1}
\end{table*}

Traditional machine learning relies on a massive amount of well-labeled centralized data and assumes that all data collected are homogeneous \cite{niu2020decade}. However, many real-world scenarios require more scalable, private, and dynamic machine learning frameworks that are capable of managing big real-time data from a variety of IoT devices. As a result, TL, FL, and OL were proposed to solve these modern challenges.

Although TL is not frequently studied as a mechanism for knowledge transmission in a decentralized environment, when combined with FL, i.e., FTL, it is capable of transmitting knowledge across distributed devices. Additionally, TL in non-federated contexts typically involves instance transmission \cite{deca31, deca32, deca33, deca34}, posing a risk of privacy leakage. FL, on the other hand, preserves privacy \cite{mcmahan2018learning, yu2020salvaging} by sharing local model update parameters instead of raw instances from local clients \cite{li2020federated}. TL enhances target model performance by providing learners in target domains a baseline performance rather than starting from scratch, thereby reducing computation overhead \cite{torrey2010transfer}. On the other hand, standard FL involves tens of millions or even billions of local devices \cite{bonawitz2019towards}, and all of these devices must meet eligibility computation power to participate in training, which is not practical as demonstrated in \cite{yang2018applied}. As a result, it is logical to apply TL to this framework in order to enable FL with clients with limited processing capability.

Real-world applications necessitate that machine learning models have a deeper understanding of the heterogeneous data and develop strong resistance to varying degrees of these scenarios \cite{kairouz2019advances}. One of the most challenging topics of heterogeneous scenarios is cross-modality \cite{niu2020decade}, as it refers to situations in which the feature and/or label spaces of the source and target domains are entirely different, which is one of the primary reasons for data heterogeneity in the majority of real-world machine learning applications. The key idea in addressing this problem is to identify feature mapping functions that project the source and target feature spaces to a common latent space via matrix factorization \cite{singh2008relational} using labeled source data or co-occurrence data \cite{CM2015-8, CM2015-9, CM2015-10, CM2015-11}. TL for cross-modality commonly transfers knowledge from easily labeled source domains to an expensively labeled target domain. For instance, consider the well-known text-to-image TL \cite{CM2015-3, CM2015-7, yang2015learning}, which leverages the semantic meaning of labeled text data to improve model classification performance on sparsely annotated image data. Besides, VFL and FTL are also applicable to cross-modality scenarios. However, the former can be used only when a specific condition is satisfied, i.e., having a large set of sample IDs that overlap between the source and target domains \cite{yang2019federated}. Additionally, while TL and FTL seek to leverage knowledge from source domains in enhancing the target model performance, the ultimate goal of VFL is to assist all source and target parties in developing a `common wealth' strategy \cite{yang2019federated}. As shown in the table, TL, FL, and FTL can all be used in cross-modality scenarios, which explains why all of these strategies are well-suited for overcoming challenges associated with a lack of well-labeled data.

Along with cross-modality heterogeneity, FL is well-suited for cross-model and cross-system scenarios due to its decentralized nature. Cross-model scenarios, which are also prevalent in fundamental machine learning applications, refer to the fact that the structure of the locally trained models varies due to the diverse data usage patterns of the local clients \cite{li2019fedmd}. FL prefers to use the global model with a predefined model paradigm as the referencing information in a cross-model scenario, and clients can update their local models based on different structures \cite{t2020personalized, lin2020ensemble}. Ensemble strategies, in which multiple learners from different source domains or learning algorithms are combined with a weight assignment strategy to maximize the utility of candidate learners that have better performance in the target domain \cite{gao2008knowledge, deca34}, are frequently used to enable TL in cross-model scenarios. Furthermore, the majority of TL paradigms are based on the premise that all learners are trained in a centralized and consistent environment whereas real-world situations are more complicated than these assumptions. On the other hand, FL is applicable in these scenarios where system heterogeneity is caused by differences in the storage, compute, and battery capacities of individual client devices. \cite{xie2019asynchronous} developed an asynchronous FL framework (FedAsync) for adaptively updating the weights of local models in response to stale information, thereby elevating FL to a more effective, flexible, and scalable level.

Moreover, TL can personalize models in non-federated environments by leveraging data from source tasks to improve performance in a related target domain. However, when TL is applied to domains that are extremely unrelated to one another, the model performance of the target domain is likely to be worse than that of the source domain without transferring the source data, a concept referred to as \textit{negative transfer} \cite{CM2015-12, CM2015-15, CM2015-16}. Similar to this concept, when local clients come from highly unrelated domains or system settings, training local models in FL for these clients using a consistent scheme can reduce the ability of each local model to depict unique client characteristics \cite{FL-Hetero-44}, causing the aggregated model to perform worse than local models trained using only their own datasets, which can be recognized as a \textit{drift} problem \cite{karimireddy2020scaffold}. One of the most widely used strategies for mitigating negative transfer is to use effective selection mechanisms to determine the relatedness (also known as transferability \cite{deca76, eaton2008modeling}) between the source and target domains prior to the transfer \cite{deca42, CM2015-19, yang2015learning}. On the other hand, the drift problem is more complicated and can be approached differently. Rather than avoiding it, the majority of researchers have chosen to turn this issue into a feature \cite{kairouz2019advances}: they create \textit{personalized} or \textit{device-specific} local models \cite{li2020federated} for clients that are intended to behave better than the aggregated global model by applying various techniques such as multi-task learning to the FL framework \cite{wang2019federated, smith2017federated, hanzely2020federated}.

One of the most prominent application scenarios in the new era of big data is modeling real-time data, which typically becomes obsolete within hours or even minutes \cite{mishne2013fast}, such as recommendation systems for business websites \cite{xiao2018personalized} and real-time non-intrusive load monitoring systems for living-alone seniors \cite{alcala2017assessing}. Additionally, there is a \textit{cold start} \cite{li2020federated} problem in real-world machine learning applications, which refers to new clients in the FL framework or new incoming datasets in the source domain in the TL system. The majority of existing TL and FL methods are based on pre-given datasets, wasting bandwidth and computational resources by requiring the framework to be retrained to achieve optimal results in the scenarios above \cite{OFL4}. Thus, it is vital to incorporate TL and FL into the OL paradigm to overcome these constraints. However, since this field is still in its infancy, few solutions have been proposed in recent years, and no prior research has summarized the research area comprehensively. After comprehending the relationship between related techniques, the following sections will provide detailed descriptions and summaries of current OTL and OFL studies to fill this review gap.

\section{Online transfer learning}\label{3}
\begin{figure}[!htbp]
\centering
\includegraphics[width=0.75\textwidth]{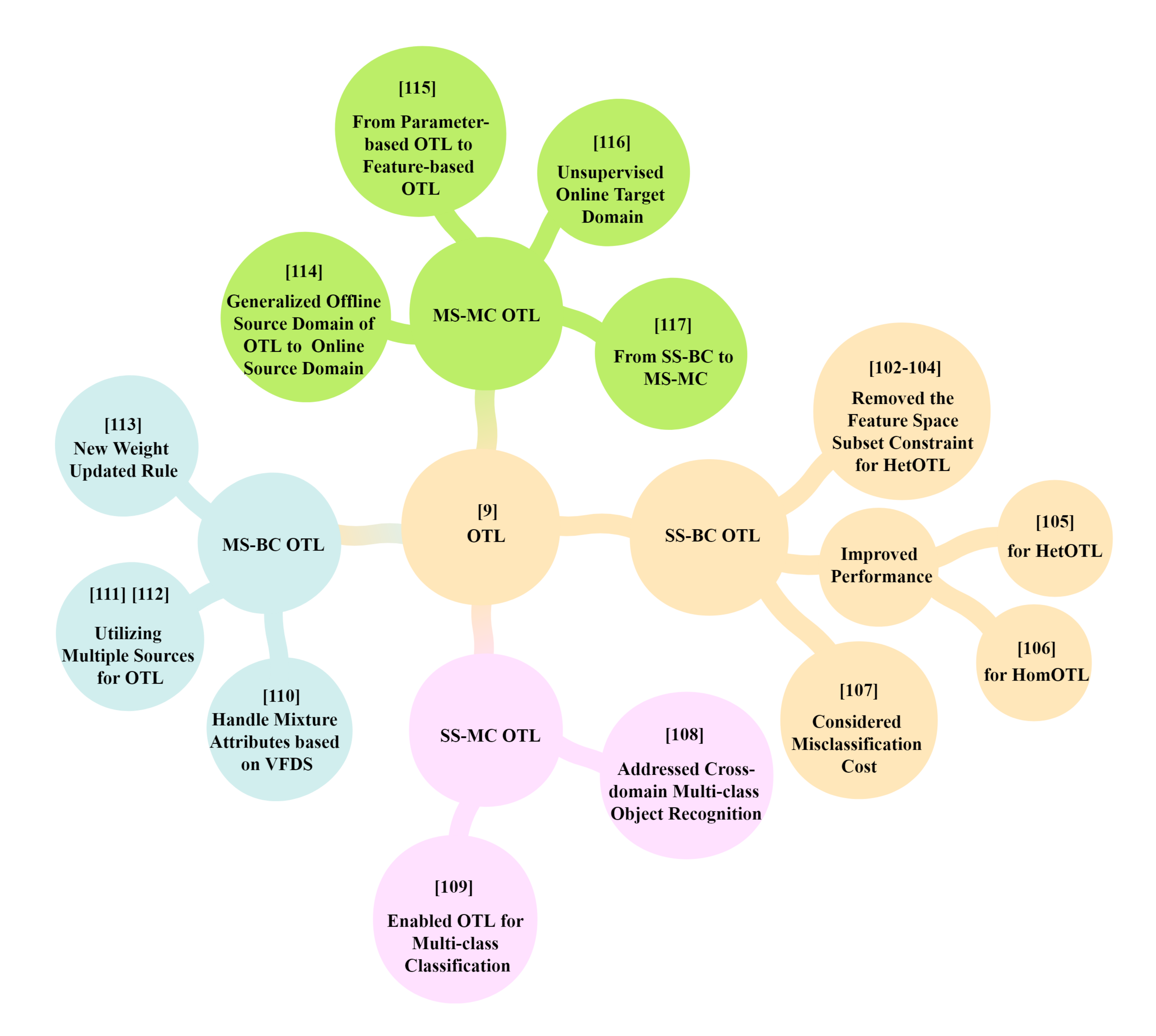}
\caption{Relation map for OTL. SS: Single source; BC: Binary classification; MS: Multiple sources; MC: Multi-class classification.}
\label{Fig7}
\end{figure}

OTL enables the standard TL paradigms to transfer knowledge from source domains, thereby enhancing the online learning task on the target domain \cite{OTL10}. Fig. \ref{Fig7} gives a relation map of OTL studies in recent years.

It is worth noting that the organization of OTL in this survey deviates from the aforementioned traditional TL categories, as it is a developing field with research focusing on a more fundamental and specific perspective. The following sections provide an interpretation of OTL approaches from a domain-task perspective. In general, domain-based interpretation is based on different settings within the source domain, including single source (SS) OTL and multiple sources (MS) OTL. On the other hand, the task-based interpretation is based on different task types within the target domain, including binary classification (BC) OTL and multi-class classification (MC) OTL. While the majority of OTL research has concentrated on classification tasks, similar techniques can be applied to other machine learning tasks such as regression and clustering \cite{OTL9, OTL10, OTL20}. According to the relation map, most existing OTL research have focused on SS-BC and MS-BC OTL while studies for SS-MC and MS-MC OTL have been relatively scarce.

\subsection{Notations and problem definition}
Table \ref{tab2} summarizes the frequently used mathematical notations in OTL, and we keep these notations consistent and similar to the majority of existing works \cite{OTL4, OTL5, OTL8, OTL10, OTL15, OTL12} to facilitate comparisons of different OTL methods.

\begin{table}[!htbp]
\centering
\caption{Summary of frequently used mathematical notations in OTL}
\begin{tabular}{cc}
\hline
\textbf{Notation}                                               & \textbf{Description}                                                                                                                                    \\ \hline
$D^{S_i}$                                                & the $i$-th source domain                                                                       \\
$n^{S_i}$                                                & the number of instances in the $i$-th source domain                                            \\
$n, K$                                                   & the number of the source domains/classes                                                     \\
$D^S$                                                    & the set of the source domains $D^{S} = \left \{ D^{S_i} \right \}^n_{i=1}$                     \\
$X^{S_i}$                                                & the feature space of the $i$-th source domain $X^{S_i} = \mathbb{R}^{d_i}$                     \\
$\mathcal{X}^T$                                          & the feature space of the target domain $\mathcal{X}^T = \mathbb{R}^{d_T}$                      \\
$\mathcal{Y}^{S_i}$                                      & the label space of the $i$-th source domain $\mathcal{Y}^{S_i}=\left \{ 1,2,\dots,k \right \}$ \\
$\mathcal{Y}^T$                                          & the label space of the target domain $\mathcal{Y}^T=\left \{ 1,2,\dots,k \right \}$            \\
$D^T$                                                    & the set of target domain                                    \\
$n^T$                                                    & the number of instances in $D^T$ \\
$\left ( x^t,y^t \right )$                               & the $t$-th arrived instance in the target domain                                               \\
$f^{S_i}(\cdot)$                                         & the model learned from the $i$-th source domain                                                \\
$f^{T}(\cdot)$                                           & the model learned from the target domain                                                       \\
$f^{t}(\cdot)$                                           & the target model                                                                               \\
$\mu_{t,i}$                                              & the weight of the $i$-th source classifier at time point $t$                                   \\
$v_{t,i}$                                                & the weight of the $i$-th target classifier at time point $t$                                   \\
$n_c$                                                    & the number of the co-occurrence instances                                                      \\
$\left( \widetilde{x}^{S_i},\widetilde{x}^{T_i} \right)$ & the $i$-th unlabeled co-occurrence data                                                               \\ \hline
\end{tabular}
\label{tab2}
\end{table}

Given $n$ source domains denoted by $D^{S}=\left \{ D^{S_i} \right \}^n_{i=1}$, where each source domain $D^{S_i}$ contains $n^{S_i}$ labeled instances. The problem of OTL is formulated with single source (SS) when $n=1$, and with multiple sources (MS) when $n>1$. The source data space in the $i$-th source domain is denote by $\mathcal{X}^{S_i}\times\mathcal{Y}^{S_i}$, where the feature space $\mathcal{X}^{S_i} = \mathbb{R}^{d_i}$.
The target domain is denoted by $D^T$, with $n^T$ instances. Similarly, we denote by $\mathcal{X}^{T}\times\mathcal{Y}^{T}$ the target data space in the target domain, where the feature space $\mathcal{X}^{T} = \mathbb{R}^{d_T}$. The problem of OTL is formulated with binary classification (BC) task when $k=2$, and with multi-class classification (MC) task when $k>2$.
When $\mathcal{X}^{S_i}=\mathcal{X}^{T}$ and $\mathcal{Y}^{S_i}=\mathcal{Y}^{T}$, the problem is identified as homogeneous OTL (HomOTL). On the other hand, if the source and target domain have different feature spaces ( $\mathcal{X}^{S_i}\ne\mathcal{X}^{T}$) or different label spaces ($\mathcal{Y}^{S_i}\ne\mathcal{Y}^{T}$), the problem is referred to heterogeneous OTL (HetOTL) \cite{pan2013transfer, OTL10}.


\subsection{Single source-binary classification (SS-BC) OTL}
\begin{figure}[!htbp]
\centering
\includegraphics[width=0.5\textwidth]{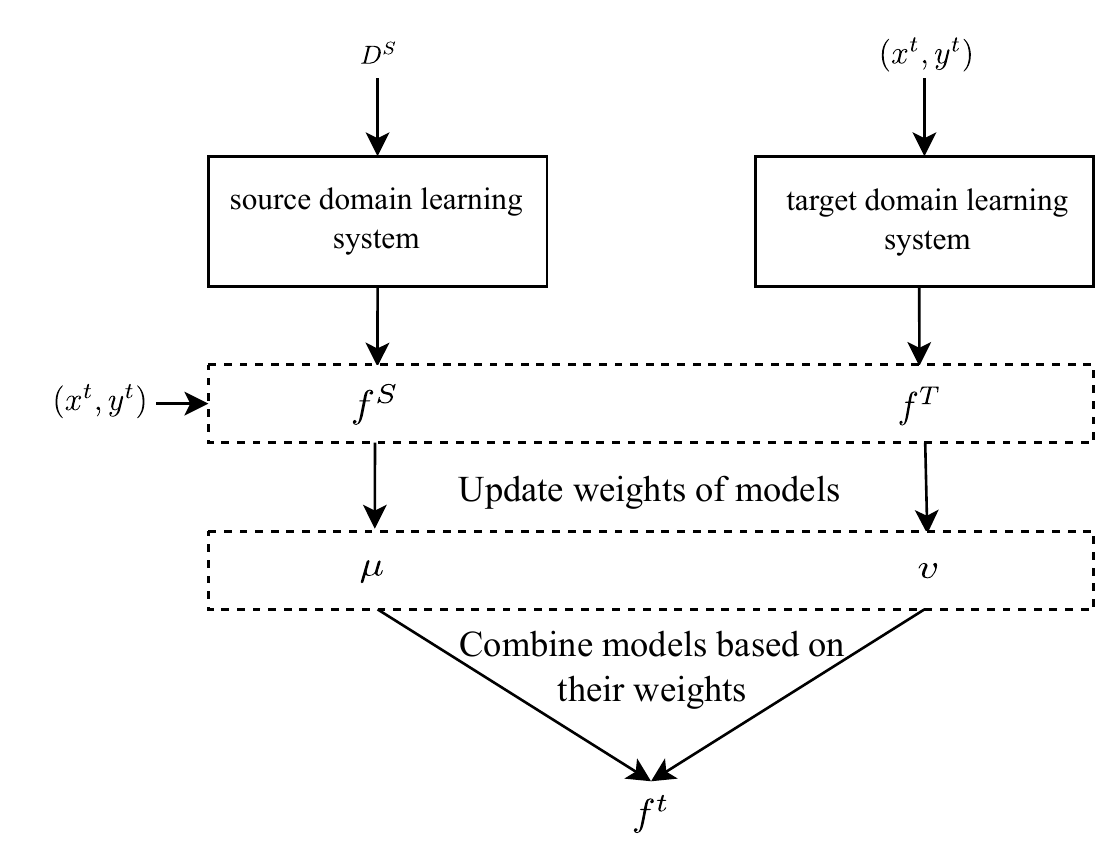}
\caption{SS-BC homogeneous OTL framework.}
\label{Fig8}
\end{figure}

SS-BC OTL was first proposed by \cite{OTL10}, which was considered in both homogeneous and heterogeneous scenarios (HomOTL and HetOTL). For HomOTL, as illustrated in Fig. \ref{Fig8}, they first constructed the source model $f^S$ using the offline source data by support vector machine (SVM) and utilized Passive-Aggressive (PA) algorithm to build model $f^T$ on the target domain. The PA formulated OL as a constrained convex optimization problem, and the weight $\omega$ of the online model on the target domain at a new time point $t+1$ was updated by the solution:
\begin{equation}
  \omega_{t+1} = \omega_t + \tau_ty_tx_t
\end{equation}
where $\tau_t= \min\left \{ \mathcal{C},\frac{\ell((x_t,y_t);\omega_t)}{\left \|x_t  \right \| }  \right \} ^2$, and $\mathcal{C}$ is a positive regularization parameter. $\ell(\cdot)$ is the hinge loss, which can be written as $\ell((x,y);\omega)=\max\left \{ 1-y(\omega^\top x),0 \right \} $. The resulting algorithm is \textit{passive}
and no update is needed when $\ell(\cdot) = 0$. Otherwise, when $\ell(\cdot)$ is positive, the algorithm is \textit{aggressive} and the instance $x_t$ will be selected as a support vector into the support vector set, which will be then forced to learn $\omega_{t+1}$. PA standardized the trade-off between progress achieved at each new time point and information gathered from prior rounds \cite{crammer2006online}.

After obtaining both the source and the target models, \cite{OTL10} proposed a weight updating scheme to adjust the weight $\mu$ of the source model and $v$ of the target model, respectively:
\begin{equation}
  \left\{\begin{matrix}
  \mu_{t+1}= \frac{\mu_ts(f^S(x_t), y_t)}{\mu_ts(f^S(x_t), y_t) + v_ts(f^T(x_t),y_t)} \hfill  \\
  v_{t+1}= \frac{v_ts(f^T(x_t),y_t)}{\mu_ts(f^S(x_t), y_t) + v_ts(f^T(x_t),y_t)} \hfill  \\
  \mu_1=v_1=\frac{1}{2} \hfill
  \end{matrix}\right.
\end{equation}
where $\mu_{t+1}$ and $v_{t+1}$ are the weights of the source and target models, respectively, at time point $t+1$. $s(\cdot)$ is a weight decay function that increases the weights of models that contribute significantly to the final forecast.

\begin{figure}[!htbp]
\centering
\includegraphics[width=0.65\textwidth]{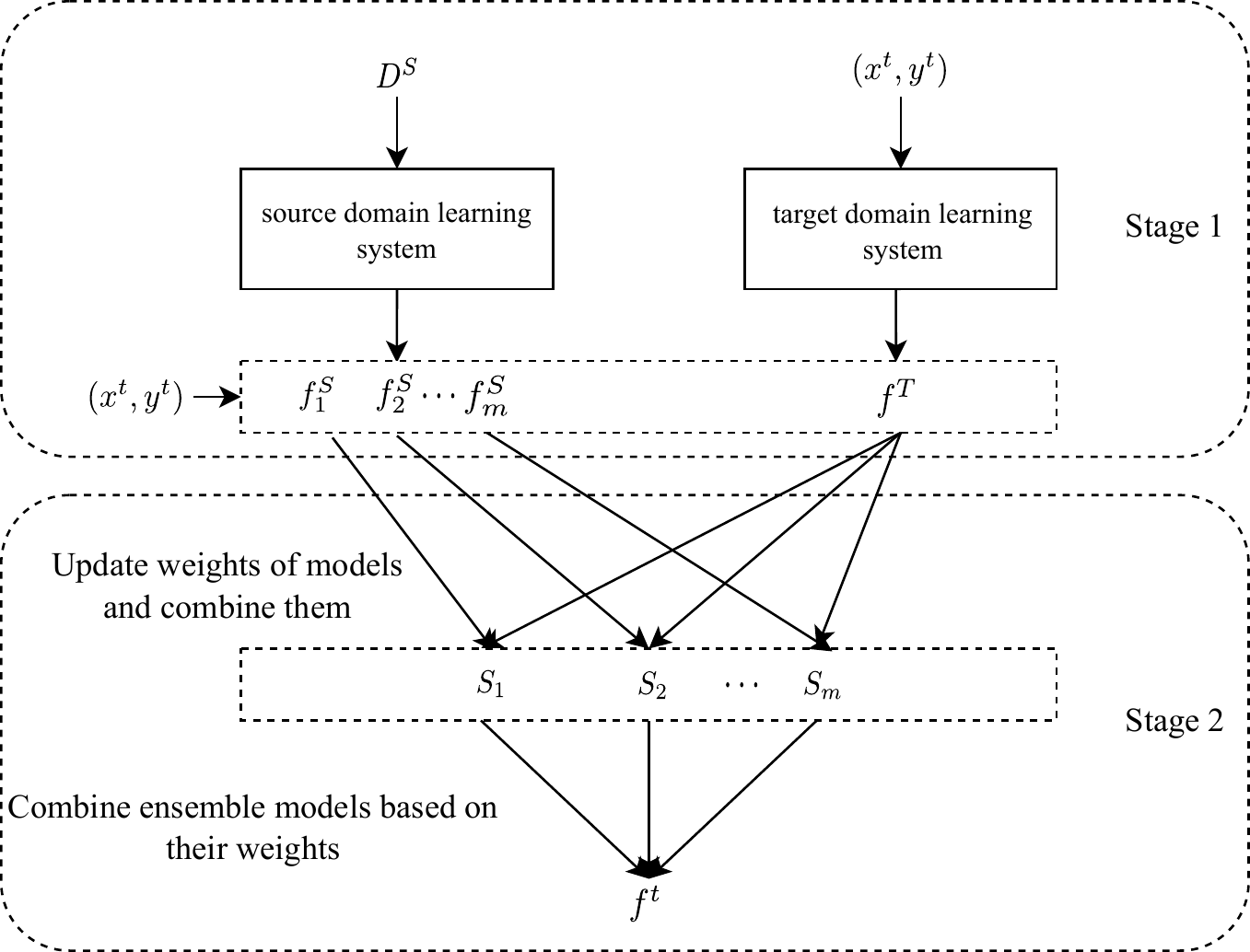}
\caption{AB-HomOTL framework.}
\label{Fig9}
\end{figure}

Unlike \cite{OTL10}, which only used a single source classifier, \cite{OTL6} proposed an AB-HomOTL inspired by boosting algorithm to learn multiple weak source classifiers. As illustrated in Fig. \ref{Fig9}, this paper focused on the learning strategy of the source model $f^S$ in the homogeneous scenario for SS-BC OTL.
Specifically, AB-HomOTL selected PA as the primary learning algorithm for training $m$ multiple weak source classifiers in the AdaBoost algorithm at the first stage. At the second stage, the source classifiers were integrated with the model $f^T$ trained on the target domain. During this stage, a weight was assigned to each combination model based on its performance on the new instance $(x^t,y^t)$. Finally, the ensemble models were integrated to produce the final robust target classifier $f^t$.

Rather than weighting classifiers dynamically according to their forecast accuracy, \cite{OTL19} emphasized that data in the real world are cost-sensitive and considered the misclassification cost to present an OTL algorithm with adaptive cost (OLAC). Specifically, they utilized the proportion of minority and majority samples to calculate the misclassification cost, enabling dynamic classifier adjustment for different samples. OLAC has been proven to be effective in improving the classification accuracy of minority samples, thereby increasing overall model performance.

\cite{OTL10} also considered the SS-BC OTL in the heterogeneous environment (HetOTL), which assumed that the feature spaces of the source domain are a subset of that of the target domain. Given a new arrived instance $(x^t,y^t)$, HetOTL divided it into two instances $(x^{t(1)}, y^t)$ and $(x^{t(2)}, y^t)$ where $x^{t(1)} \in \mathcal{X}^S$ and $x^{t(2)} \in \mathcal{X}^T / \mathcal{X}^S$. Then, inspired by multi-view approaches, HetOTL trained and updated two classifiers $f^{T(1)}$ and $f^{T(2)}$ from two views simultaneously using co-regularization optimization:
\begin{equation}
  \begin{split}
      (f^{T(1)}_{t+1}, f^{T(2)}_{t+1}) = \mathop{\arg\min}_{f^{T(1)},f^{T(2)}}\frac{\gamma_1}{2}\left \| f^{T(1)}-f^{T(1)}_t \right \|^2 \\ +\frac{\gamma_2}{2}\left \| f^{T(2)}-f^{T(2)}_t \right \|^2+\mathcal{C}\ell( f^{T(1)}, f^{T(2)};t)
  \end{split}
\end{equation}
where $\gamma_1$, $\gamma_2$ and $\mathcal{C}$ are predefined positive regularization parameters, and $\ell(\cdot)$ is the loss function. During the updating, the classifier $f^{T(1)}_1$ was initialized by the trained source classifier $f^S$, and the other classifier $f^{T(2)}_1$ was initialized to 0. This updating rule ensured that the two-view classifiers did not deviate excessively from the previous updates (first two regularization terms) while maintaining the prediction performance (the last term).

Similar to \cite{OTL6}, \cite{OTL9} proposed heterogeneous ensembled OTL (HetEOTL) based on AdaBoost to improve the performance of OTL models in a heterogeneous environment. The comparative experiment demonstrated that employing the ensemble strategy outperformed the previous HetOTL framework in \cite{OTL10}. Although \cite{OTL9} improved the performance of the OTL model, it made the same assumption as \cite{OTL10}, i.e., the feature spaces of the source domain are a subset of that in the target domain.

\begin{figure}[!htbp]
\centering
\includegraphics[width=0.8\textwidth]{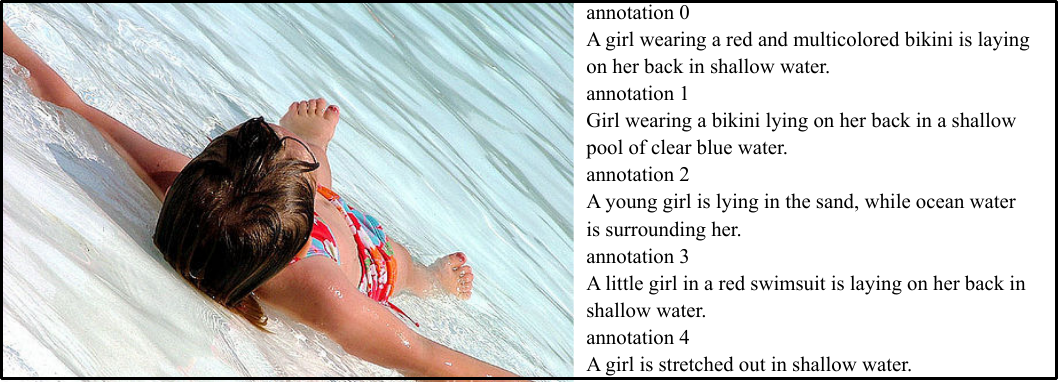}
\caption{An instance of co-occurrence text-image data from Flicker \cite{young2014image}}
\label{Fig10}
\end{figure}

To relax the above assumption, studies based on co-occurrence data have been proposed \cite{OTL4, OTL7, OTL21}. Given a source domain $D^S$ and a target domain $D^T$ where the feature spaces of them are totally diverse, i.e., $\mathcal{X}^S \cap \mathcal{X}^T = \emptyset$. The unlabeled co-occurrence data $\left\{ \left( \widetilde{x}^{S_i}, \widetilde{x}^{T_i} \right) \right\}^{n_c}_{i=1} \in \mathcal{X}^S\times\mathcal{X}^T$ are collected from offline sources to bridge different feature spaces, in which $\widetilde{x}^{S_i} \in \mathcal{X}^S$ and $\widetilde{x}^{T_i} \in \mathcal{X}^T$. For example, the website Flickr\footnote{http://www.flickr.com} contains a massive collection of images with tags that can be used as co-occurrence data and are significantly less expensive to collect than labeled image data (Fig. \ref{Fig10}).

\begin{figure}[!htbp]
\centering
\includegraphics[width=0.5\textwidth]{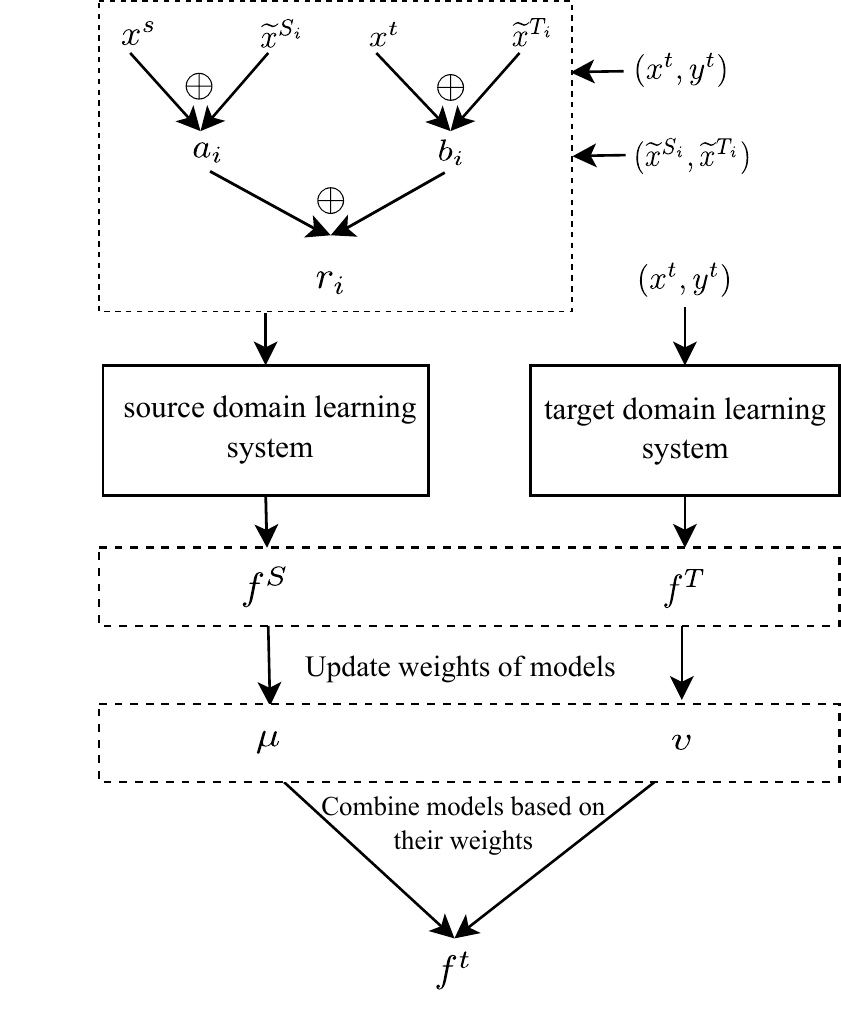}
\caption{OHTHE framework. The $\oplus$ maker denotes the measure of similarity between two instances.}
\label{Fig11}
\end{figure}

\cite{OTL21} proposed online heterogeneous transfer learning by hedge ensemble (OHTHE), which utilized co-occurrence data as auxiliary knowledge to build a correspondence map between the source and target domains, as illustrated in Fig. \ref{Fig11}.
 They first measured the heterogeneous similarity between the newly arrived instance $x^t$ and the offline source instance $x^s$ based on co-occurrence text-image data. The source model was then built by adding the weights of the $k$ nearest neighbors of $x^t$ in the source domain. Meanwhile, the model on the target domain was trained by PA. Following that, the OHTHE utilized the Hedge ($\beta$) strategy \cite{freund1997decision} to updated the weights $\mu$ and $v$ dynamically:
\begin{equation}
    \left\{\begin{matrix}
    \mu_{t+1}= \mu_t\beta^{\ell(y_tf^S(x_t))}  \hfill  \\
    v_{t+1}= v_t\beta^{\ell(y_tf^T(x_t))} \hfill  \\
    \mu_1+v_1=1  \hfill
    \end{matrix}\right.
  \end{equation}
where $\mu_1 \in (0,1)$ and $v_1 \in (0,1)$ are the initial weights. $\beta$ is a weight decay factor that is used to identify models that contribute more to the final prediction and whose magnitude is determined by the loss function $\ell(\cdot)$.

\subsection{Multiple sources-binary classification (MS-BC) OTL}
In real-world applications, it is difficult to extract sufficient knowledge from a single source domain, and combining information from multiple source domains makes the source classifiers more reliable and robust. However, directly combining all source domains may result in unsatisfactory forecasts since different source domains include information from different perspectives, and the data qualities within each source domain vary as well. As a result, OTL algorithms with multiple sources should be more sophisticated in order to distinguish critical source domains and thus construct a more robust source learner.

\cite{OTL18} trained a set of source classifiers using the kernel SVM, and each classifier was weighted according to its performance on the newly arrived instance of the target domain. The weighted source classifiers were then integrated to create an ensemble learner for the source domain. Simultaneously, PA was used to train the target classifier on the online target data. The ensemble source and target classifier were then integrated to generate an effective ensemble model at the second stage. The weight updating rule at the next round $t+1$ of the classifier from $i$-th source domain, the ensemble source classifier, and the target classifier can be described as follows:
 \begin{equation}
\left\{\begin{matrix}
    \mu_{t+1}^i= \mu_t^i\beta^{(f^{S_i}(x_t), y_t)} \hfill  \\
    \mu_{t+1}= \mu_t\beta^{(f^S(x_t), y_t)} \hfill  \\
    v_{t+1}= v_t\beta^{(f^T(x_t),y_t)} \hfill  \\
    \mu_1^i=\frac{1}{2n} \hfill \\
    \mu_1=v_1=\frac{1}{2} \hfill
\end{matrix}\right.
\end{equation}
where $f^S={\textstyle \sum_{i=1}^{n}}\mu_t^if^{S_i} (x_t) $. $\beta \in (0,1)$ is a weight decay factor that is applied when the classifier suffers a loss value, and $\mu^i_t$ denotes the weight of the classifier from the $i$-th source domain at time point $t$.

In contrast to \cite{OTL18}, which only investigated HomOTL, \cite{OTL15} adapted the OTL framework to a heterogeneous environment. Similar to the problem setting in \cite{OTL10}, \cite{OTL15} introduced heterogeneous OTL with multiple source domains (HetOTLMS), which was based on the premise that feature spaces of the source domain are a subset of that of the target domain. Instead of training an ensemble source classifier, HetOTLMS combined the weak classifier from $i$-th source domain with the target classifiers trained by PA to form $n$ ensemble classifiers. In particular, for the $i$-th source domain in the $t$-th round, each newly arrived instance was divided into two pieces, the first of which shared the same feature space as the source domain, and the second of which shared the remainder of the target feature space. Two classifiers
on the target domain were generated and then integrated with the source classifier
based on their weights to form an ensemble classifier.

The majority of studies developed models based on PA that were limited to numerical attributes. Inspired by the very fast decision tree (VFDT), which incorporates Hoeffding bounds to guarantee the performance of an incremental decision tree, \cite{OTL16} modified VFDT as VFDT-D in the following ways to provide an OTL framework capable of dealing with mixed attributes:
  \begin{itemize}
    \item Cache a few instances to initialize the statistical information for newly constructed leaf nodes to satisfy the Hoeffding constraint and manage mixture attributes.
    \item Modify the output form of the VFDT to treat it as a posteriori probability equal to the ratio of positive training instances in a leaf node with respect to the total number of training instances in that leaf node.
  \end{itemize}

Then, using the VFDT-D, decision trees were induced from the source domains and the target domain. Following that, the tree path and posterior probability of the newly arrived instance $x_t$ were then combined to determine the ideal source domain with the highest degree of similarity to $x_t$, which was then integrated with the target domain classifier to construct the final prediction decision function.
Comparative experiments demonstrated that the proposed algorithm was capable of overcoming the cold start problem \cite{li2020federated}, which occurs when the model performance degrades in the early stage of the data stream due to the low number of instances arriving in the target domain.

It is worth noting that the target model performs worse than the source model as it lacks prior knowledge about the target domain. As more instances arrive, the target model will perform equally well or even better than the source model. On the other hand, most studies \cite{OTL10, OTL18, OTL21} updated model weights solely based on cumulative error, ignoring the intrinsic timescale of online data. To address this issue, \cite{OTL12} proposed a new weight updating rule that assigns a higher weight to later occurrences. They assumed that the predictions made by the newer samples were more plausible than those made by the earlier samples and hence increased the weights over time to narrow the gap between the accuracy and the weights of the models. While the traditional accumulating criteria ensure that outliers have a negligible effect on the models, examining whether the same scenario holds in this framework is necessary.

\subsection{Single source multi-class classification (SS-MC) OTL and multiple sources multi-class classification (MS-MC) OTL}
Following the discussion of the binary classification OTL frameworks in the previous section, we will discuss multi-class classification OTL studies in this section. Numerous tasks in the real world, such as document classification, are multi-class. Specifically, when an instance is relevant to a single subject, the classification problem is referred to as multi-class single-label classification; otherwise, the classification problem is referred to as multi-class multi-label classification \cite{crammer2006online}, and the majority of existing OTL research has focused on multi-class single-label classification. Multi-class classification is more complicated than binary classification as it involves the development of offline and online models that take multiple classes, necessitating the use of more sophisticated strategies to create a combined multi-class classifier with satisfactory performance \cite{OTL21}.

Inspired by the online multi-class PA (MPA) algorithm \cite{crammer2006online}, \cite{OTL2} presented an OTL algorithm for multi-class classification (OTLAMC) that adopted a novel loss function and weight update mechanism to enable OTL in multi-class classification tasks. However, this paper only concentrated on knowledge transfer from a single source domain. \cite{OTL5} then developed the online multi-source transfer learning for multi-class classification (OMTL-MC) system, which incorporated data from multiple domains. While the OMTL-MC structure is similar to that of the HetOTLMS framework described in \cite{OTL15}, there are two significant differences:
\begin{itemize}
  \item The OMTL-MC framework examined OTL in a homogeneous environment, whereas the HetOTLMS framework investigated OTL in both homogeneous and heterogeneous settings.
  \item OMT-MC was developed with an extended Hinge loss (EHL) function to support multi-class classification tasks whereas HetOTLMS is only suitable for binary classification.
\end{itemize}

\cite{OTL22} proposed an online PA feature transformation (OPAFT) algorithm to calculate the similarity in a $k$ nearest neighbor ($k$-NN) classifier. Furthermore, they extended this algorithm to the online multiple kernel feature transformation (OMKFT) algorithm to improve the performance of OPAFT for cross-domain and multi-class object recognition. Another feature-based OTL framework was proposed in \cite{OTL8}, which investigated multi-class classification OTL with multiple source domains. Specifically, they constructed an initial transformation matrix for the $i$-th source domain by utilizing source and target data. Then, the transformation matrix was used to project the original data onto a new feature space. Meanwhile, the newly arrived instance was projected to its appropriate feature space using all of the transformation matrices, and a new source classifier was trained in this new space. The projected instance was then trained using the MPA algorithm to generate the associated classifiers for the target domain. Finally, the source and target classifiers were combined using the Hedge strategy. Rather than updating the transformation matrices at each time step, this paper used a time window to control the frequency of updates, thereby reducing the computing cost.

In contrast to previous OTL architectures that required label revealing of target instances after each prediction, \cite{OTL14} introduced an online multiple source transfer learning (OMS-TL) architecture that requires only a few labeled data points in the target domain as a priori and does not require label revealing after each prediction. They employed a bipartite graph to represent the classification results from all the source domains and then estimated the likelihood of a sample belonging to each class using convex minimization. When a new instance is observed, the averaged probability from all source domain classes to which the sample belonged was combined with the target prediction based on the weighted average of previous predictions to generate the final result.

OTL aims to enhance the online learning task on the target domain by leveraging knowledge from source domains. By applying standard TL in the online context, real-time data generated by various edge devices can be processed efficiently. However, as with traditional TL, OTL is constrained by the assumption that all data from the source and the target domains must be processed centrally, which is impractical in the real world due to data privacy regulations. As a result, the following section will introduce OFL, which enables real-time data processing in a distributed way while maintaining data privacy.

\section{Online federated learning}\label{4}
Standard FL has been constrained to the premise that the training data at each local device is gathered offline and should be fully trained throughout each global round to deliver iteration round-efficient solutions \cite{OFL1}. On the other hand, FL holds significant promise for a variety of sophisticated applications, including smart traffic management \cite{liu2020privacy}, interactive social networks \cite{mishne2013fast}, and smart health monitoring \cite{brisimi2018federated}, owing to the massive amounts of data generated by various edge devices (e.g., smartphones, wireless sensors, and wearable devices). It is impossible to assume that the training data at each local client remains constant throughout each round of training, as clients may have access to real-time data that will become obsolete in a matter of hours or even minutes \cite{mishne2013fast, bajwa2020machine}. In this case, standard FL models will have difficulty capturing the fluctuations of real-time data, and their generalization performance is likely to decrease with an increase in training rounds. Therefore, enabling the standard FL architecture in online scenarios (i.e. OFL) is critical in the big data era. Instead of delivering iteration round-efficient solutions by simply waiting for training results from all the local clients, OFL studies are increasingly focusing on the real-time data processing efficiency of local clients, i.e., on delivering iteration process-efficient solutions \cite{OFL1}.

OFL considers that the data from each client is generated and collected in real-time, and it seeks to capture a high degree of temporal information from various distributions of data sources. Due to the time-varying nature of online data, several of the challenges associated with standard FL are becoming increasingly apparent in the online FL:
\begin{itemize}
  \item Statistical heterogeneity: non-IID and unbalanced properties of online time-varying data cause model/concept drift \cite{OFL11} in OFL, and capturing the dynamic change of the rapidly generated online data is a significant challenge for OFL.
  \item System heterogeneity: stragglers emerge due to device heterogeneity and network unreliability. Balancing the contribution of each local device to the local iteration against the communication cost of the global iteration is a critical challenge in OFL.
  \item Privacy guarantees: with the massive amount of online data generated, providing privacy guarantees for OFL becomes more difficult. Various privacy protection strategies, such as differential privacy (DP) \cite{dwork2014algorithmic}, have been implemented in FL in order to strike a balance between data utility and privacy, and these techniques should be optimized for the online environment to be more reasonable and practical for OFL.
\end{itemize}

Different OFL research focuses on different challenge priorities, and TABLE \ref{Tab3} summarizes current OFL studies on those three challenges. The table demonstrated that the majority of OFL studies focused on statistical and system heterogeneity, while more research is needed in the area of providing privacy guarantees for OFL.


\begin{table}[!htbp]
\caption{Summary of studies on OFL}
\centering
\begin{tabular}{llll}
\hline
\multicolumn{4}{l}{\textbf{Online Federated Learning}}                          \\ \hline
Statistical Heterogeneity      & \multicolumn{3}{l}{ \cite{OFL4, OFL11, OFL2, OFL7, OFL13, OFL19}}     \\
System Heterogeneity        & \multicolumn{3}{l}{\cite{OFL1, OFL2, OFL19, OFL3, OFL6, OFL8, OFL9}} \\
Privacy Guarantees              & \multicolumn{3}{l}{\cite{OFL12, OFL14}}                  \\ \hline
\end{tabular}\label{Tab3}
\end{table}

\subsection{Notations and problem definition}
Assume that we have a set of $\mathcal{K} =\left\{1 \ldots K \right\}$ distributed devices. At each round $t$, the global server broadcasts its most recent parameters $\omega_g^t$ to the $K$ devices. Each local device $k$ receives the global parameters $\omega_g^t$ and a time-varying local instance $(x_k^t,y_k^t)$ to update the parameters $\omega_k^{t+1}$ of its local model $f_k(\omega_k^{t+1})$. Finally, the local devices upload the updated parameters $\omega_k^{t+1}$ to the global server for dynamic aggregation:
\begin{equation}
  \omega_g^{t+1} = h(\omega_1^{t+1}\ldots \omega_K^{t+1})
\end{equation}
mapping $h$ should be carefully selected in accordance with the model parameter structures \cite{OFL8}, from which each device $k$ can estimate the label $y_k^{t+1}$ of a newly arrived data $x_t^{k+1}$ in real-time. One of the most commonly used mappings in the standard FL system is \textit{FedAvg} \cite{mcmahan2017communication}, which averages the aggregated local parameter sets:
\begin{equation}
\omega_g^{t+1} = \sum_{k=1}^{K}\frac{n_k}{N}\omega_k^{t+1}
\end{equation}
where $n_k$ is the number of the data samples taken on device $k$, and $N$ is the total number of samples taken on $K$ local devices. In OFL scenarios, the data is generated continuously on local devices, increasing the uncertainty of local models in comparison to the central model \cite{OFL2}. As a result, more plausible mappings should be established to constrain such variances and thus improve the generalization performance of the model. Additionally, as not all devices are activated during each round $t$ for a variety of reasons (e.g. due to network delay or device heterogeneity), strategies such as devices selection should be used to mitigate the negative effect on overall communication efficiency.

\subsection{OFL with statistical heterogeneity}

\cite{OFL13} concentrated on the statistical heterogeneity associated with unbalanced data in OFL. Specifically, \cite{OFL13} assumed that the central server had been provided with pre-given data for training the initial central model. After initialization, the central model was broadcast to local devices to be trained with new samples from different classes. Then, the updated models on local devices were uploaded to the central server for integration. Then, the integrated model was optionally retrained using pre-given training data in the central server to ensure that it did not deviate significantly from the original central model. This strategy effectively addressed common OL challenges, such as the catastrophic forgetting \cite{shoham2019overcoming} that occurs due to the time-varying nature of online data.

To enable OFL framework in non-IID scenarios, \cite{OFL7} designed a non-linear regression OFL framework based on random Fourier feature-based kernel least-mean-square (RFF-KLMS). Specifically, they defined a non-linearly local model $f_k(\omega_k^t)$ for an arrived instance $(x_k^t, y_k^t)$ of local device $k$ at time $t$. Then the local parameter updating function can be formulated as:
\begin{equation}
\omega_k^{t+1} = \mathbb{E} \left [ 	\vert  f_k(\omega_k^t) -\hat{f_k}(\omega_k^t)	\vert \right ]
\end{equation}
where $\mathbb{E}\left [\cdot \right]$ is the expectation. $ \hat{f_k}(\omega_k^t) = \omega_k^Tz_k^t$, where $\omega_k$ is a linear representation of the non-linear model $f_k$ in the random Fourier feature (RFF) space, and $z_k^t$ is the mapping of $x_k^t$ in the RFF space. Hence, the global parameter updating function can be further constructed in the RFF space by:
\begin{equation}
\omega_g^{t+1} = \frac{1}{K}\sum_{k=1}^{K} \omega_k^{t+1}.
\end{equation}

Instead of training a common utility global model for all local devices, some works have concentrated on improving the performance of personalized local models in the OFL framework.
Based on the worker-leader-core network hierarchy, researchers designed hierarchical nested personalized federated learning (HN-PFL) for unmanned aerial vehicles (UAVs) \cite{OFL11}. The intra-UVA swarm is nested within an inter-UVA aggregation, which follows the worker-leader-core network structure to train high-level personalized models for local devices. To enhance the learning of HN-PFL, \textit{model/concept drift} was introduced to quantify the dynamic change of local online time-varying data. For a local device $k$ with its local model $f(\omega_k^t)$, denote the online model drift at time $t$ by $\Lambda_k^t\in\mathbb{R^+}$, which could capture the upper bound of the variation of parameters between two adjacent instances, we have:
\begin{equation}
\left \| \bigtriangledown f_k(\omega_k^{t+1})- \bigtriangledown f_k(\omega_k^{t}) \right \|^2\le \Lambda _k^{t+1}.
\end{equation}
Local models with a greater drift value imply that they will become obsolete more easily, necessitating a shorter learning period and requiring revisiting more frequently. On the other hand, models with a small drift value have lower local parameter fluctuation, implying that they require less attention than models with greater drift values. Then, the model drift value was utilized to estimate the online gradient of each local model and created a core network for each training sequence by storing the real-time properties of the network as reinforcement learning states. Additionally, to avoid the curse of dimensionality, they used a neural network to model the Q-table and determine the network states, rather than pre-building it using traditional reinforcement learning techniques \cite{mnih2015human}.

\cite{OFL4} emphasized the importance of developing individualized local models by combining multi-task learning with the OFL framework. Unlike previous works that analyzed streaming data, \cite{OFL4} proposed an online federated multi-task learning framework (OFMTL) to address the problem of inferring effective local models for newly joined devices without affecting previous clients or the global server. The multi-task relationship learning \cite{zhang2010convex} was used in the OFMTL to transfer the relationship between the local models of all the devices into a relationship precision matrix. The OFMTL formulated the learning of model parameters of the newly joined device as a convex optimization problem relating to the weight matrix and the precision matrix, and an alternating optimization algorithm was proposed for alternatively optimizing the model parameters and precision matrix of the new device using the information gained from previous devices. Additionally, to save computation resources while retaining the generalization performance of previous models, the model parameters were configured to be retrained only when the number of newly joined devices reached a fixed ratio with respect to the total number of previous devices.

\subsection{OFL with system heterogeneity}
The varying communication rate of the heterogeneous local device is another critical challenge for OFL, and lagging devices with a lower communication rate in this system are known as \textit{stragglers} \cite{OFL2}. Numerous solutions have since been proposed for standard FL systems. However, in real-world scenarios where data on each local device fluctuates, the updated model for each global round may exhibit more inherent dynamic characteristics. Therefore, more sophisticated algorithms for FL in the online context (i.e. OFL) need to be designed to mitigate the negative impact of stragglers in a dynamic environment. Based on the reviewed papers, two types of protocols can be used to address the issue of stragglers: (1) synchronous protocol; (2) asynchronous protocol.

To deal with stragglers in the OFL system, \cite{OFL1} proposed an adaptive batch sizing (ABS) solution using the synchronous protocol. Typical synchronous FL systems require the central server to wait for all local devices (including stragglers) to be updated before performing a global update \cite{jiang2017heterogeneity, zhang2016staleness}, or simply ignores and drops the stragglers \cite{mcmahan2017communication}. Different from the above studies, ABS \cite{OFL1} limited the size of training data at each global round by allocating a batch size bound to each device based on its processing speed and real-time data generation speed, forcing all local devices, including stragglers, to be synchronous during each global communication round. Furthermore, ABS provided a buffer for each local device to keep or revisit local data depending on network settings to reduce the volatility of produced data size during each training round. Despite a lack of mathematical definitions in \cite{OFL1}, the proposed ABS structure in this paper is instructive.

\cite{OFL19} proposed a cost-effective federated learning (CEFL) system capable of cooperatively reducing computation and communication overhead. Similar to \cite{OFL1}, CEFL made dynamic decisions on local devices by limiting the entry of newly arrived training data, buffering, and scheduling the data according to the time-varying resource pricing of local devices. Additionally, CEFL employed an additional optimization parameter
to balance the computational overhead of local models and the overall communication cost.

\begin{figure}[!htbp]
\centering
\includegraphics[width=0.7\textwidth]{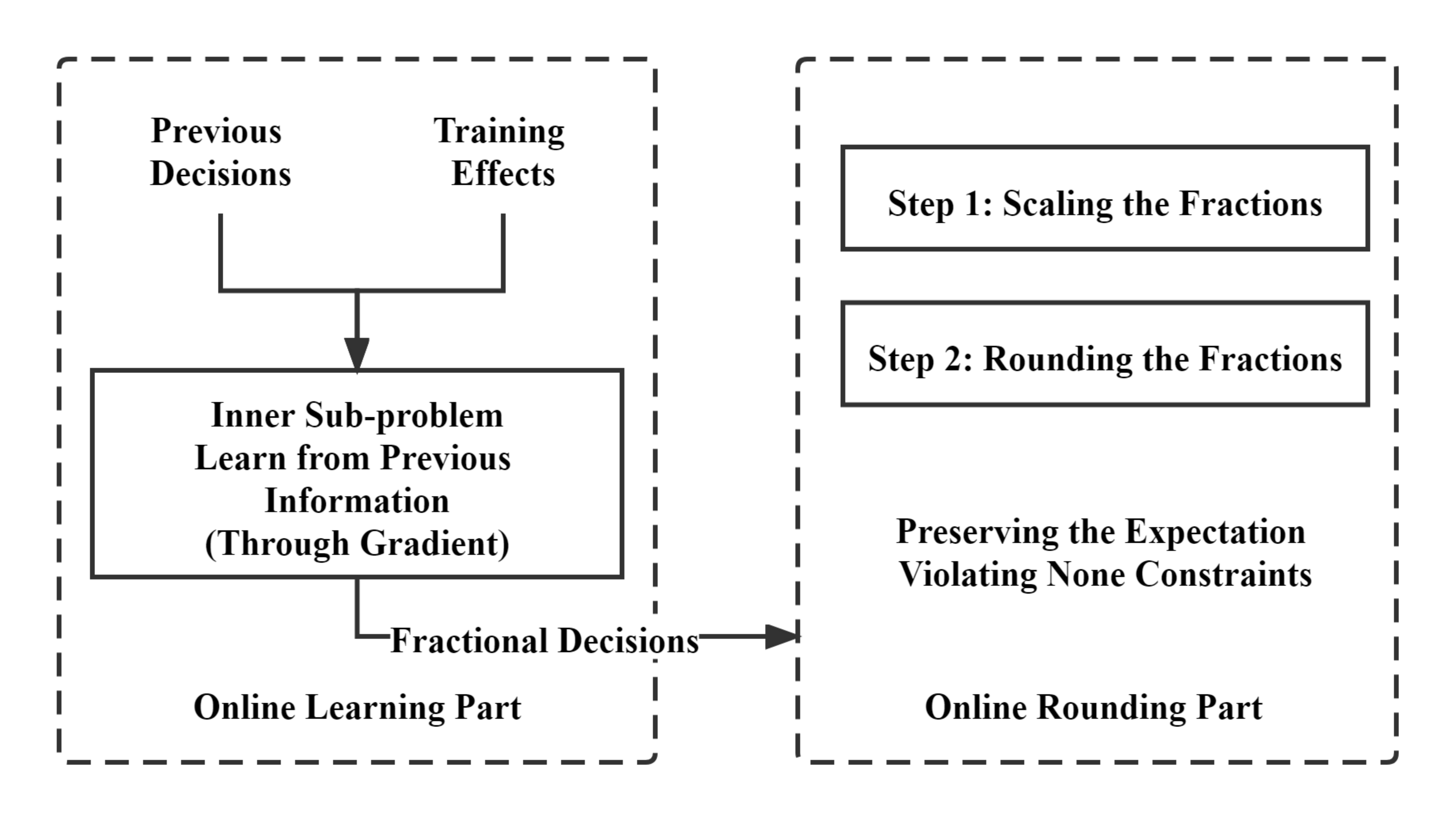}
\caption{Structure of the online schema proposed in \cite{OFL9}.}
\label{Fig12}
\end{figure}

Rather than leveraging all local devices for global model training, \cite{OFL9} highlighted that the key challenge for OFL is to distinguish between effective local models and to determine the appropriate number of local epochs without any future knowledge. This study formulated the participant selection problem as an optimization problem based on system capacity (local device availability, data volume, and network bandwidth) and long-term convergence of both local and aggregated global models. To further extend this solution into online scenarios, an online schema was designed (Fig. \ref{Fig12}) to dynamically select the participant and the number of local epochs for each device in the OFL system. Specifically, the online schema consists of two parts: online learning and online rounding. The first part produced fractional judgments solely based on prior knowledge, whereas the second part employed a compensation technique to randomly convert the fractional decisions to integers without breaking any pre-defined constraints. The experiment results indicated that the proposed schema could dynamically adjust the upper bound on local convergence accuracy and select participants with superior local model performance and computation efficiency.

Asynchronous system design was also emphasized in studies addressing the heterogeneity of the OFL system.

\cite{OFL2} developed an asynchronous OFL framework (ASO-Fed) that enabled a wait-free OFL system and improved the prediction performance and computation efficiency of local devices when data arrived continuously. ASO-Fed learned the inter-client interaction on the global server using feature representation learning inspired by attention mechanisms \cite{firat2017multi, vaswani2017attention} and weight normalization \cite{wang2006normalization, ba2016layer}. The decay coefficient was utilized on the client-side to execute OL on each local device, balancing the older and newer models. Additionally, ASO-Fed used a dynamic step size to mitigate the negative impact of stragglers. The step size was determined by the data volume and communication capacity of each client, and a larger step size was assigned to local clients with a lower activation rate to compensate for the long latency and achieve higher performance. Experiments demonstrated that the proposed ASO-Fed framework converged more quickly than synchronized FL frameworks and significantly reduced overall computational overhead.

On the other hand, \cite{OFL6} highlighted the importance of incorporating contributions from all local customers, even stragglers. They developed FL{\small EET}, which consists of two components, I-P{\small ROF} and A{\small DA}SGD. The first component aims to forecast and allocate computation overhead across all local devices. The latter is a novel stochastic gradient descent algorithm that employs weighted stale gradients determined by the stale-aware dampening factor and the similarity-based boosting value. Stragglers with longer delays were assigned a smaller stale-aware dampening factor, indicating that they contribute less to the overall updating process. In contrast, a lower similarity value represented a more significant gradient containing more valuable new data features. The FL{\small EET}
has been proved to be effective in mitigating the negative effects of stragglers while also capturing vital information to improve the generalization capability of the system.

\subsection{OFL with privacy guarantees}
Providing privacy guarantees for the OFL framework necessitates a more sophisticated design of the privacy algorithms since the data are generated in an online fashion and the entire training data sequence is unknown. \cite{OFL12} considered P2P FL in an online setting, and proposed an online mirror descent algorithm with long-term constraints on the sequential decisions made by each device. Additionally, a modified online version of local differential privacy was utilized to ensure the privacy of the OFL system. Instead of relying on loss information from the entire data sequence to provide global privacy guarantees, the online local differential privacy required only the private version of loss gradients for real-time data sequence at each global round. At each global training round, each user received new data and updated its local model. After that, each updated local gradient was subjected to local differential privacy to ensure privacy in the online scenarios. When compared to the online gradient descent algorithm with differential privacy, the proposed algorithm was proved to be more accurate in the long run.

In large-scale online distributed network settings, the dynamic growth of the online dataset complicates the process of incorporating noise into each associated data sequence to ensure privacy. \cite{OFL14} utilized a trusted third party to protect the privacy of OFL in a recommendation system based on adaptive binary tree-based noise aggregation. They constructed a binary item-cluster tree for each local device to reduce the scale of incoming online big data at each global round. Specifically, the item space was partitioned into refined child clusters, and the optimal recommendation for the corresponding client was searched from top to bottom of the constructed tree. Then, to ensure privacy, a trusted third party was proposed as a middleware to provide safe model aggregation over all agents using an exponential mechanism, and two forms of attacks from internal local devices and external adversaries were evaluated.

OFL enables real-time distributed data training while maintaining data privacy, and much of the current literature on OFL focuses on addressing statistical and system heterogeneity rather than providing privacy guarantees. Compared with OTL studies, there are relatively few historical studies in the area of OFL. Based on the discussion of the methodology in the above two sections, we will explore OTL and OFL from practical aspects in the next section to give a more comprehensive description of these methods.

\section{Practical aspects in online federated and transfer learning}\label{5}
Although studies of OTL and OFL have been conducted with promising results in a variety of fields in recent years, there are still practical concerns that need to be addressed. This section discusses the practical issues associated with online federated and transfer learning from two perspectives: datasets and applications.

\subsection{Datasets}
We introduce several datasets frequently used in OTL and OFL and then discuss practical considerations and concerns.

\subsubsection{Popular datasets for OTL}
Four datasets have been commonly used in OTL: \textit{20Newsgroups} dataset, \textit{sentiment analysis} dataset, \textit{text-image} dataset, and \textit{Office-Caltech} dataset. The \textit{20Newsgroups} dataset\footnote{http://qwone.com/\url{~}jason/20Newsgroups/} contains about 20,000 newsgroup documents organized by subject and subcategory. The \textit{20Newsgroups} dataset has been mainly used to implement MS-BC OTL tasks \cite{OTL12, OTL15, OTL18}. Typically, researchers focus on two primary subjects, each of which has multiple subtopics. Then, to simulate multiple learning domains, a positive label is assigned to each subtopic, which corresponds to the negative label assigned to a subtopic within the other primary subject.

Another commonly used dataset is the \textit{sentiment analysis} dataset\footnote{http://www.cse.ust.hk/TL/index.html}, which consists of product reviews on Amazon for four different product domains (books, DVDs, electronics, and kitchen). Each review includes a human rating score (0-5 stars), a review caption, position, timestamp, an item description, a reviewer name, and the review content. This dataset has been used to perform SS-BC OTL \cite{OTL6, OTL9, OTL10}
and MS-BC \cite{OTL15, OTL18} OTL tasks.

The \textit{Office-Caltech} dataset \cite{gong2012geodesic} is made up of real-world object domains gathered from the Berkeley Office \cite{saenko2010adapting} and Caltech-256\footnote{https://resolver.caltech.edu/CaltechAUTHORS:CNS-TR-2007-001}, which has been widely utilized in OTL tasks requiring multi-class classification. The Caltech-256 has 30,607 pictures from 256 groups, and the real-world object domains include Amazon, Webcam, and the digital single-lens reflex camera.

Different from the above three datasets, the \textit{text-image} dataset has been utilized in a wide range of cross-modality OTL scenarios, and it is sourced from the NUS-WIDE \cite{chua2009nus} collection on Flickr. This dataset comprises photos and tags that have been published on the internet and is often used in heterogeneous SS-BC OTL. More precisely, the unlabeled text-image data pairs in this dataset are often utilized as co-occurrence data to bridge the text samples from the source domain and the images from the target domain.

\subsubsection{Popular datasets for OFL}
Several datasets have been popularly utilized in OFL: \textit{CIFAR-10 and CIFAR-100}, \textit{MINIST}, and \textit{air quality} dataset. \textit{MINIST} and \textit{CIFAR} are two public datasets that are often utilized in OFL tasks \cite{OFL2, OFL6}, particularly for simulations of non-IID settings.

\textit{CIFAR-10 and CIFAR-100} datasets\footnote{https://www.cs.toronto.edu/\url{~}kriz/cifar.html} both have 60,000 images, with the former having 10 classes with 6,000 photos each and the latter having 100 classes with 600 images each.

\textit{MINIST}\footnote{http://yann.lecun.com/exdb/mnist/} is a database of handwritten digits that cotains a training set of 60,000 instances and a testing set of 10,000 instances. This dataset is suitable for pattern recognition tasks as it requires minimal pre-processing.

\textit{Air quality} datasets collected from weather sensors in different countries were used in \cite{OFL2} and \cite{OFL8} to predict the level of pollutants in the air.

\subsubsection{Practical considerations}
OTL tasks have primarily been conducted on public datasets such as \textit{Office-Caltech}, which may have storage format limitations and are prone to becoming obsolete. It is also challenging and difficult to update these existing datasets or re-collect fresh datasets. On the other hand, real-world datasets are difficult to obtain due to privacy regulations. Furthermore, most datasets only include a limited number of labeled instances in the target domains, making it challenging to perform cross-validation to fine-tune the target model \cite{OTL22}. For example, OTL applications for the healthcare system are primarily derived from publicly released hospital data, and these applications may be limited to patients in a particular geographical area, as people in different regions may have varying physical conditions. Additionally, an OTL system may require target patients to upload their physical states in near real-time, which is highly unlikely in practice due to privacy concerns and system/ infrastructure limitations.

When designing a comparative experiment, different domain types of OTL tasks require different data settings and must comply with the same data dividing rule. On the other hand, data settings for OFL are relatively complex, which requires to simulate both the statistical heterogeneity generated by non-IID or imbalanced data and the system heterogeneity caused by varying uploading rates across numerous local devices in an online scenario. To deal with statistical heterogeneity, researchers often use the standard data decentralization method \cite{bonawitz2019towards} to classify the data and partition the data in each category into multiple shards of varying sizes, after which each local client is allocated with different shards \cite{OFL2, OFL6}. For the simulation of stragglers, a random delay timer can be used to reflect various network delays across local clients \cite{OFL2}. Furthermore, a data growth rate should be predetermined to imitate the growth of online data. Data settings for OFL involve a variety of parameters, and it is important to establish a unified standard for these parameters to facilitate comparative experiments.

\subsection{Applications}
Although research and practice in OTL and OFL are still in their infancy, several studies have identified major prospects for them and sparked a series of related investigations and efforts to apply them to real-world problems. This section will describe the cutting-edge applications and discuss relevant practical considerations. It
covers the application scenarios and their compatibility of both OTL and OFL in different contexts, with the recognition that existing studies can be categorized into two sectors of industrial engineering and healthcare.

\subsubsection{Applications in industrial engineering}
Given the achievements of OTL in domain shift scenarios and OFL in data privacy protection, it is reasonable to apply these methods to industrial engineering tasks, and Table \ref{Tab4} summarizes the detailed sub-scenarios of OTL and OFL applications in industrial engineering.

\begin{table}[!htbp]
\caption{Summary of sub-scenarios of OTL and OFL in industrial engineering}
\centering
\begin{tabular}{lll}
\hline
\textbf{Sub-Scenarios}                & OTL                                    & OFL                     \\ \hline
Environmental Protection     & $\backslash$                            & \cite{OFL2, OFL8}       \\
Unmanned Aerial Vehicle      & $\backslash$                             & \cite{OFL10, OFL11}     \\
Sentiment Analysis           & \cite{OTL6, OTL9, OTL10, OTL15, OTL18} & $\backslash$              \\
Image Recognition            & \cite{OTL5, OTL8, OTL20, OTL22}        & \cite{OFL2, OFL3, OFL6} \\
Online Recommendation Systems & \cite{OTL26}                           & \cite{OFL14}            \\ \hline
\end{tabular}\label{Tab4}
\end{table}

OFL has been used in a variety of data-sensitive industry domains, including environmental protection \cite{OFL2, OFL8}, and unmanned aerial vehicle (UAV) control \cite{OFL10, OFL11}. OTL applications, on the other hand, frequently appeared in industrial situations involving domain shift problems, such as sentiment analysis \cite{OTL6, OTL9, OTL10, OTL15, OTL18}. There are other situations in the industrial engineering where data is likely to be sensitive and therefore a cross-domain task is required, such as image recognition \cite{OTL5, OTL8, OTL20, OTL22, OFL3, OFL2, OFL6} and online recommendation systems \cite{OFL14, OTL26}.

\cite{OFL2} forecast the pollutants in the air by combining data from multiple weather sensors located in nine separate locations to develop the best collaborative OFL model. Apart from the environmental protection, OFL has been used to control UAVs in real-time \cite{OFL10, OFL11} to support many mission-critical applications such as first-aid packet dispatching and firefighting \cite{ackerman2018medical, tisdale2009autonomous}.

Sentiment analysis has arisen as a hot topic in OTL, with applications ranging from spam detection \cite{OTL14} to document categorization \cite{OTL9, OTL12, OTL18}. For example, \cite{OTL14} developed a spam email filter system by analyzing real-world emails from fifteen different users. Such a system can help reduce labor costs whilst safeguarding the property of people.

Image recognition is a trending topic in OTL. Transferring information from related domains makes it possible to conduct online image classification on the target domain. In addition, computer vision-based tasks \cite{OFL3} are a hot topic in industrial applications of OFL. Rather than uploading annotated personal visual data to a central database, participants in an object recognition task can train a local model on their personal site. Furthermore, by leveraging an online learning framework, OFL enables computer vision-based tasks to manage massive amounts of online image data that arrive sequentially from cameras.

Additionally, OFL and OTL have been implemented in online recommendation systems. \cite{OTL26} proposed SocialTransfer, a cross-domain OTL system for multimedia applications that learns from time-varying social stream data. To address the privacy concerns associated with information sharing, \cite{OFL14} developed a privacy-preserving recommendation system based on OFL that takes advantage of the privacy guarantees provided by the federated learning architecture while still capable of managing the streaming data.

\subsubsection{Applications in healthcare}

Both OFL and OTL are promising solutions for healthcare. For OFL, the connection between real-time data monitoring from various edge devices and hospital records breaks down analysis barriers between various parties while maintaining data privacy. Furthermore, the information required to detect a disease differs from patient to patient. Given that the medical records of each patient constitute a unique domain, OTL is well-suited for disease diagnosis, as it can leverage multiple patient records to improve the diagnosis accuracy of the target patients. OTL has been used to diagnose a wide variety of diseases, including arrhythmias \cite{OTL19}, breast cancer \cite{OTL19}, and epileptic seizures \cite{OTL17}.

Nowadays, with the rapid development in storage capacity and computing power of edge devices such as smartphones and wearable devices (e.g., google glass), physical data about daily human life can be collected and analyzed conveniently. These data, however, are sensitive and are at risk of being compromised by unauthorized access. On the other hand, real-time monitoring systems are required for special scenarios, such as remote health condition monitoring for the elderly living alone, as certain acute-onset diseases (e.g., heart attack, stroke) must be detected instantly. With privacy guarantees, OFL is an excellent candidate for the aforementioned application scenarios, and it has been used in a variety of healthcare applications, including human activity recognition \cite{OFL2, OFL4}, and eating habits monitoring \cite{OFL4}.

\subsubsection{Practical considerations}
Existing research on OTL has primarily concentrated on text/image-based applications, which might become unusable in some scenarios involving users who are not familiar with text/image inputting. There are studies on TL that have recommended the use of more forms of inputs, such as voices \cite{devlin2019bert} and gestures \cite{cote2019deep}. Future OTL research should consider extending these advanced applications to online contexts, which would accommodate a variety of inputs and facilitate human-machine interaction.

While current application domains of OTL and OFL are primarily focused on industrial engineering and healthcare, there are numerous application areas worth exploring in TL and FL, such as smart transportation \cite{liang2019federated}. Traditional offline frameworks for smart transportation are likely to benefit from an online environment; for example, establishing an online autonomous driving system can capture the dynamic nature of the vehicle system and the inherent uncertainty in real-life environment, aiding drivers to make more accurate and timely decisions.

With the widespread use of edge devices, device owners can annotate their data simply by tagging or labeling on the device, which has been frequently utilized in OFL research. On the other hand, malicious and false tagging will become more prevalent with tagging on the local devices by local users becoming possible. OFL will need to concentrate on filtering the invalid tagging to ensure the accuracy of model inferences in the future. Moreover, fewer OTL applications have utilized smart edge devices as a result of personal data privacy regulations. We anticipate that the performance of OTL models trained on real-time data generated by edge devices will be significantly improved. Therefore, it is anticipated there will be  future research opportunities to combine OTL and OFL to develop an online FTL framework that takes advantage of both OTL and OFL paradigms to accomplish this vision. After investigating OTL and OFL from practical perspectives, we will conclude this survey and discuss several future works worthy of consideration in the following section. In particular, we will present a vision of online FTL and describe the proposed framework in detail.

\section{Discussion and Conclusion}\label{6}

\renewcommand\arraystretch{1.3}
\begin{table*}[!htb]
\caption{Frontier implementation scenarios of different techniques}
\resizebox{\textwidth}{13mm}{
\begin{tabular}{lccclcccc}
\hline
\multicolumn{1}{c}{\multirow{2}{*}{}} & \multirow{2}{*}{\textbf{Decentralization}} & \multicolumn{3}{c}{\textbf{Heterogeneity}}                                                                     & \multirow{2}{*}{\textbf{\begin{tabular}[c]{@{}c@{}}Inadequate Well-labeled \\ Data\end{tabular}}}
& \multirow{2}{*}{\textbf{Privacy-Preserving}} &
 \multirow{2}{*}{\textbf{\begin{tabular}[c]{@{}c@{}}Client-Side \\ Personalization\end{tabular}}}
 & \multirow{2}{*}{\textbf{\begin{tabular}[c]{@{}c@{}}Real-time \\ Data\end{tabular}}} \\ \cline{3-5}
\multicolumn{1}{c}{}                  &                                            & \multicolumn{1}{l}{\textbf{Cross-Modality}} & \multicolumn{1}{l}{\textbf{Cross-Model}} & \textbf{Cross-System} &                                                                                                 &                                              &                                                                                                  &                                                                                                                                                                                             \\ \hline
\textbf{Transfer Learning}            & \XSolid                                    & \Checkmark                                  & \Checkmark                               & \XSolid               & \Checkmark                                                                                      & \XSolid                                                                                       & \Checkmark                                                                                 & \XSolid                                                                                          \\
\textbf{Federated Learning}           & \Checkmark                                 & \Checkmark                                  & \Checkmark                               & \Checkmark            & \Checkmark                                                                                      & \Checkmark                                                                                     & \XSolid                                                                                 & \XSolid                                                                                             \\
\textbf{Federated Transfer Learning}  & \Checkmark                                 & \Checkmark                                  & \Checkmark                               & \Checkmark            & \Checkmark                                                                                      & \Checkmark                                                                                      & \Checkmark                                                                                 & \XSolid                                                                                          \\
\textbf{Online Transfer Learning}            & \XSolid                                    & \Checkmark                                  & \Checkmark                               & \XSolid               & \Checkmark                                                                                      & \XSolid                                                                                       & \Checkmark                                                                                 & \Checkmark                                                                                        \\
\textbf{Online Federated Learning}            & \Checkmark                                 & \Checkmark                                  & \Checkmark                               & \Checkmark            & \Checkmark                                                                                      & \Checkmark                                                                                     & \XSolid                                                                                 & \Checkmark                                                                                      \\
\hline
\textbf{Online Federated Transfer Learning}            & \Checkmark                                 & \Checkmark                                  & \Checkmark                               & \Checkmark            & \Checkmark                                                                                      & \Checkmark                                                                                     & \Checkmark                                                                                 & \Checkmark                                                                                      \\
\hline
\end{tabular}}
\label{Tab5}
\end{table*}
In this survey, we have provided a systematic and comprehensive overview of OTL and OFL. OTL employs knowledge from single or multiple source domains to train online target models for the target domain while OFL is a method enabling online models at the edge of distributed networks. We discussed the unique properties of OTL from a domain-task perspective and described existing research on OFL addressing several major challenges. Moreover, popular datasets and cutting-edge online federated and transfer learning applications were summarized, and practical considerations were presented from the perspectives of datasets and applications. In the following, we will identify open problems worthy of future research efforts, and also propose a vision of online federated transfer learning - a new framework conceptualised by us with the aim to deal with the most significant challenges faced in existing studies.

From the methodology perspective, existing OTL studies have mainly focused on SS-BC and MS-BC OTL while studies for multi-class classification OTL tasks have been relatively scarce. Therefore, sophisticated OTL frameworks for various types of learning tasks need to be developed in future research. Besides, current OTL frameworks mostly adopted the kernel method to build their online target classifiers. It has the distinct benefit of being more accurate than linear models. However, the disadvantage of being resource-intensive in terms of support vector storage is also well acknowledged. It is recommended that efficient solutions such as budget online kernel learning \cite{hoi2021online}, which restricts the number of support vectors to a fixed budget, are to be included in the future OTL framework since they have the potential to minimize computing overhead significantly. On the other hand, studies in the field of OFL frequently focused on developing effective models for a variety of asynchronous devices. Moreover, all current OFL frameworks, whether synchronous or asynchronous, have assumed that local devices are available during their allocated `working period', which is impractical because unforeseen events may occasionally occur, rendering these local devices being unavailable. As a result, a feedback mechanism could be developed in the future OFL framework to confer sufficient authority on the local device to commence the communication process.

From the practical perspective, existing OTL studies often utilize public datasets, and the real-world datasets are difficult to obtain due to data privacy regulations since OTL is based on the assumption that all models will be trained on a central device. Therefore, there is a need of collecting more state-of-the-art datasets for OTL tasks. On the other hand, OFL datasets are more diverse since the local clients can retain their datasets on the local device. However, typical OFL tasks often require complex data settings for simulating the heterogeneous scenarios in the real world, and different settings of the datasets make the comparisons between different OFL frameworks difficult. Therefore, developing unified data setting protocols is also necessary for future research. Moreover, the most prevalent learning type in real-world applications is supervised learning for OTL and OFL, which involves label-revealing after each prediction. Although significant progress has been achieved in online federated and transfer learning for handling distributed time-varying data with few labels, applications for unsupervised learning continue to be a barrier in this field. Methods such as \cite{wang2020unsupervised}, which made use of a selective pseudo-labeling strategy and achieved high performance for unsupervised TL, and federated unsupervised representation learning \cite{van2020towards}, which pre-trained deep neural networks using unlabeled data in a federated setting, have shown promising outcomes recently. FL and TL, as two forms of collaborative training, hold tremendous potential in the domain of unsupervised learning. Given the dynamic requirements of real-world machine learning, it is reasonable to argue subsequent works on FL and TL extensions for unsupervised learning in online contexts are needed.

The implementation scenarios of TL, FL, FTL, OTL, and OFL are summarized in Table \ref{Tab5}, and the ideal implementation scenarios of online FTL are also given in the table. As can be seen from Table \ref{Tab5}, OTL enables standard TL to handle real-time data efficiently. As with the standard TL, OTL is not frequently studied in a decentralized environment, and it often involves the instance transmission process, which poses a risk of data privacy violations. On the other hand, OFL can handle real-time data generated on local devices and also provide privacy guarantees. However, similar to standard FL, OFL needs to utilize special techniques such as TL to create personalized local models.
Since FTL has gained increasing attention with research having demonstrated its efficiency \cite{liu2020secure}, we envisage that extending FTL to online scenarios will enable the development of an advanced machine learning framework with dynamic natures that leverages both OTL and OFL paradigms.
\begin{figure}[!htbp]
\centering
\includegraphics[width=0.5\textwidth]{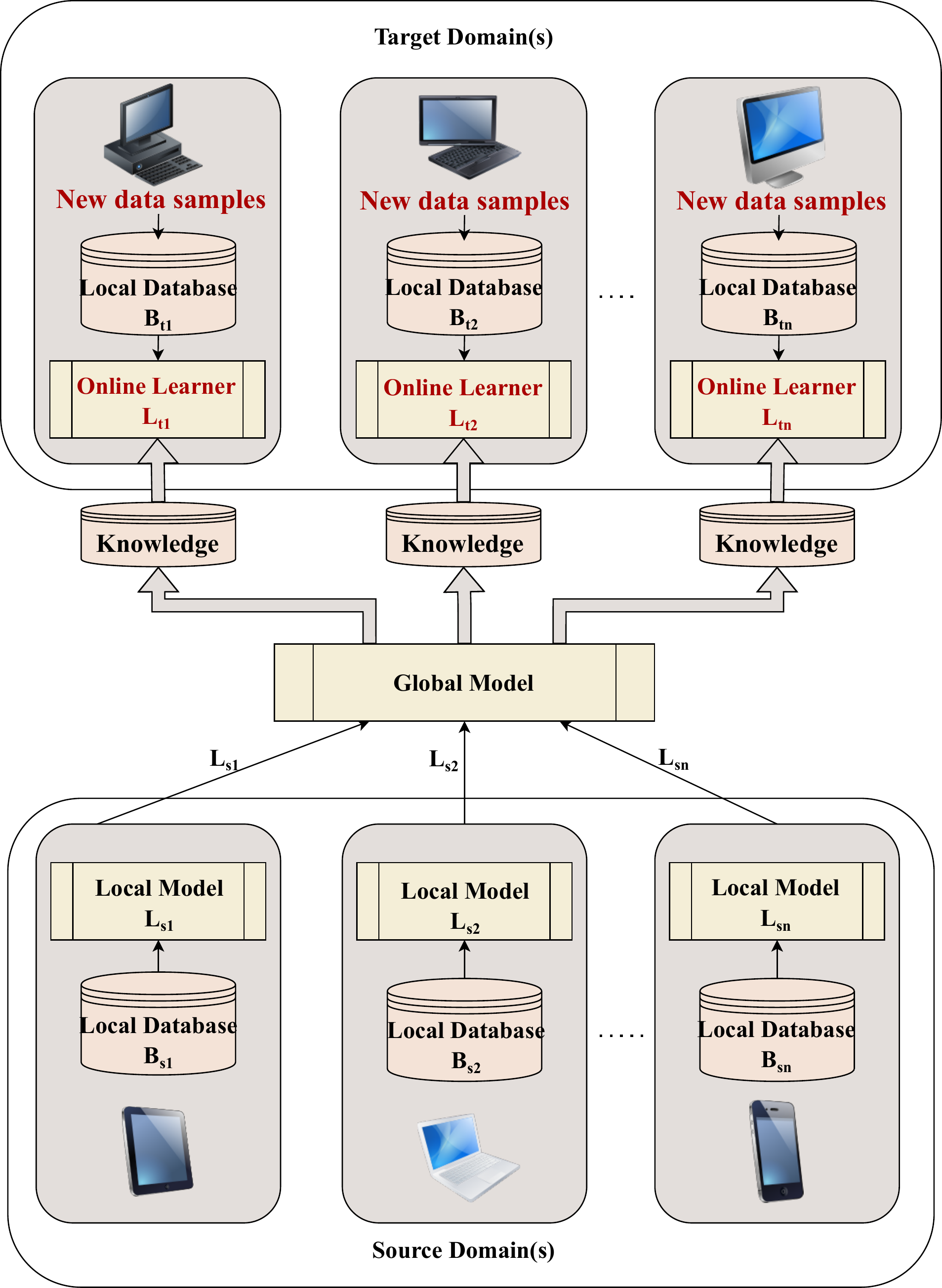}
\caption{A vision of online FTL framework}
\label{Fig13}
\end{figure}

The proposed online FTL framework is illustrated in Fig. \ref{Fig13}, and described below. The data in the source domain can be generated in real-time or from pre-given datasets. It should be noted that a scratch of the source data is essential to ensure the benchmark performance of the source models. Each local device in the target domain generates data in an online fashion, and the real-time data is analyzed by online learners, which aim to find the optimal strategy to make the online updates at each training round \cite{hoi2021online}. The global model enables model aggregation, heterogeneous computing, updating, and broadcasting. Local devices, such as smartphones and laptops, provide essential infrastructure tools, including local online/offline training, uploading, and distributed storage.

Various applications may be built on top of the proposed online FTL to provide critical human-machine interface services. By utilizing federated learning, machine learning models for multiple parties can be established without exporting local data, ensuring data security and privacy while providing users with tailored and targeted services. Meanwhile, the combination of TL enables FL to train models on a variety of different but related parties, which is practically important given that stakeholders within the same FL framework are usually from the same sector. Furthermore, classical batch/ offline learning has low efficiency in terms of computing costs, as well as limited scalability for large-scale applications due to the need of model retraining when online data sequences are generated. We envisage that extending FTL to online scenarios will help overcome the limitations of traditional batch learning by allowing online learners to update the local model rapidly and effectively.

To summarize, this survey aims to serve as a resource for researchers and practitioners developing online federated and transfer learning frameworks, by providing a systematic and comprehensive overview of OTL and OFL, and identifying open research questions worthy of future research efforts. Providing solutions to those new arising research problems from methodologies to practical applications will necessitate collaborative and long-term efforts from various research communities.

\bibliography{Pref}

\end{document}